\documentclass[11pt]{article}

\usepackage[]{acl}
\usepackage{listings}
\usepackage[most]{tcolorbox}
\usepackage{lipsum} 
\usepackage{xcolor}
\usepackage{amssymb}
\usepackage{pifont}
\lstdefinelanguage{json}{
  basicstyle=\ttfamily\small,
  numbers=left,
  numberstyle=\tiny,
  stepnumber=1,
  numbersep=5pt,
  showstringspaces=false,
  breaklines=true,
  frame=single,
  backgroundcolor=\color{gray!5},
  xleftmargin=2em,
  xrightmargin=0pt,
  framexleftmargin=1em,
  framexrightmargin=0.5em,
  framesep=.5em,
  literate=
   *{0}{{{\color{blue}0}}}{1}
    {1}{{{\color{blue}1}}}{1}
    {2}{{{\color{blue}2}}}{1}
    {3}{{{\color{blue}3}}}{1}
    {4}{{{\color{blue}4}}}{1}
    {5}{{{\color{blue}5}}}{1}
    {6}{{{\color{blue}6}}}{1}
    {7}{{{\color{blue}7}}}{1}
    {8}{{{\color{blue}8}}}{1}
    {9}{{{\color{blue}9}}}{1}
    {:}{{{\color{black}:}}}{1}
    {,}{{{\color{black},}}}{1}
    {"}{{{\color{red}"}}}{1}
}

\tcbset{
  promptbox/.style={
    colback=gray!10,
    colframe=black,
    boxrule=0.8pt,
    arc=2mm,
    outer arc=2mm,
    boxsep=5pt,
    left=5pt,
    right=5pt,
    top=5pt,
    bottom=5pt,
    fonttitle=\bfseries,
    title=Prompt
  }
}
\usepackage{times}
\usepackage{latexsym}
\usepackage{amsmath}
\usepackage[T1]{fontenc}

\usepackage[utf8]{inputenc}

\usepackage{microtype}
\usepackage{subcaption}

\usepackage{inconsolata}

\usepackage{graphicx}
\usepackage{booktabs}
\usepackage{multicol}
\usepackage{multirow}
\title{Are Large Language Models Effective Knowledge Graph Constructors?}

\author{
Ruirui Chen$^{1}$, Weifeng Jiang$^{2}$, Chengwei Qin$^{3}$, Bo Xiong$^{4}$, Kaiwen Wei$^{5}$, \\ \bf Fiona Liausvia$^{1}$, Pei Fang Tan$^{1}$, Ker Yung Chua$^{6}$, Dongkyu Choi$^{1}$, \\ \bf Mukkesh Kumar$^{1}$, Evelyn C. Law$^{1, 6, 7}$, Dennis Wang$^{1, 8, 9}$, Boon Kiat Quek$^{1}$\\
\textsuperscript{1}Agency for Science, Technology and Research;
\textsuperscript{2}Nanyang Technological University;\\
\textsuperscript{3}Hong Kong University of Science and Technology (Guangzhou);
\textsuperscript{4}Iowa State University;\\
\textsuperscript{5}Chongqing University;
\textsuperscript{6}National University Hospital; 
\textsuperscript{7}National University of Singapore; \\
\textsuperscript{8}Imperial College London;
\textsuperscript{9}University of Sheffield
}

\begin{document}
\maketitle
\begin{abstract}
Knowledge graphs (KGs) are widely used in knowledge-intensive applications, yet it remains unclear how effectively current large language models (LLMs) can construct document-grounded KGs in a zero-shot, schema-free setting without relying on complex task-specific frameworks. We introduce Detail-to-Abstract Hierarchical Knowledge Graph (D2A-HKG) construction framework, which decomposes KG construction into three stages: initial extraction, splitting, and abstraction, and evaluates the resulting graphs from both semantic and structural perspectives. Using seven frontier LLMs, we benchmark zero-shot KG construction on CMW-Lit, a dataset derived from published paediatric research articles on children's mental well-being. CMW-Lit provides a challenging test bed due to its heterogeneous evidence, interconnected factors, and complex, statistically qualified relationships. Our results show that state-of-the-art LLMs can generally produce relevant and document-faithful triples with limited hallucination, while exhibiting substantially different extraction behaviors across the construction stages. These findings provide empirical insight into the strengths and limitations of frontier LLMs for direct knowledge graph construction. We further release CMW-Lit and the resulting knowledge graphs as resources for future research, with the generated graphs providing a strong foundation for expert refinement and downstream knowledge-intensive applications\footnote{\url{https://github.com/rrchen2026-eai/D2A-HKG}}.

\end{abstract}
\section{Introduction}
\begin{figure*}[t]
    \centering
    \includegraphics[width=2\columnwidth]{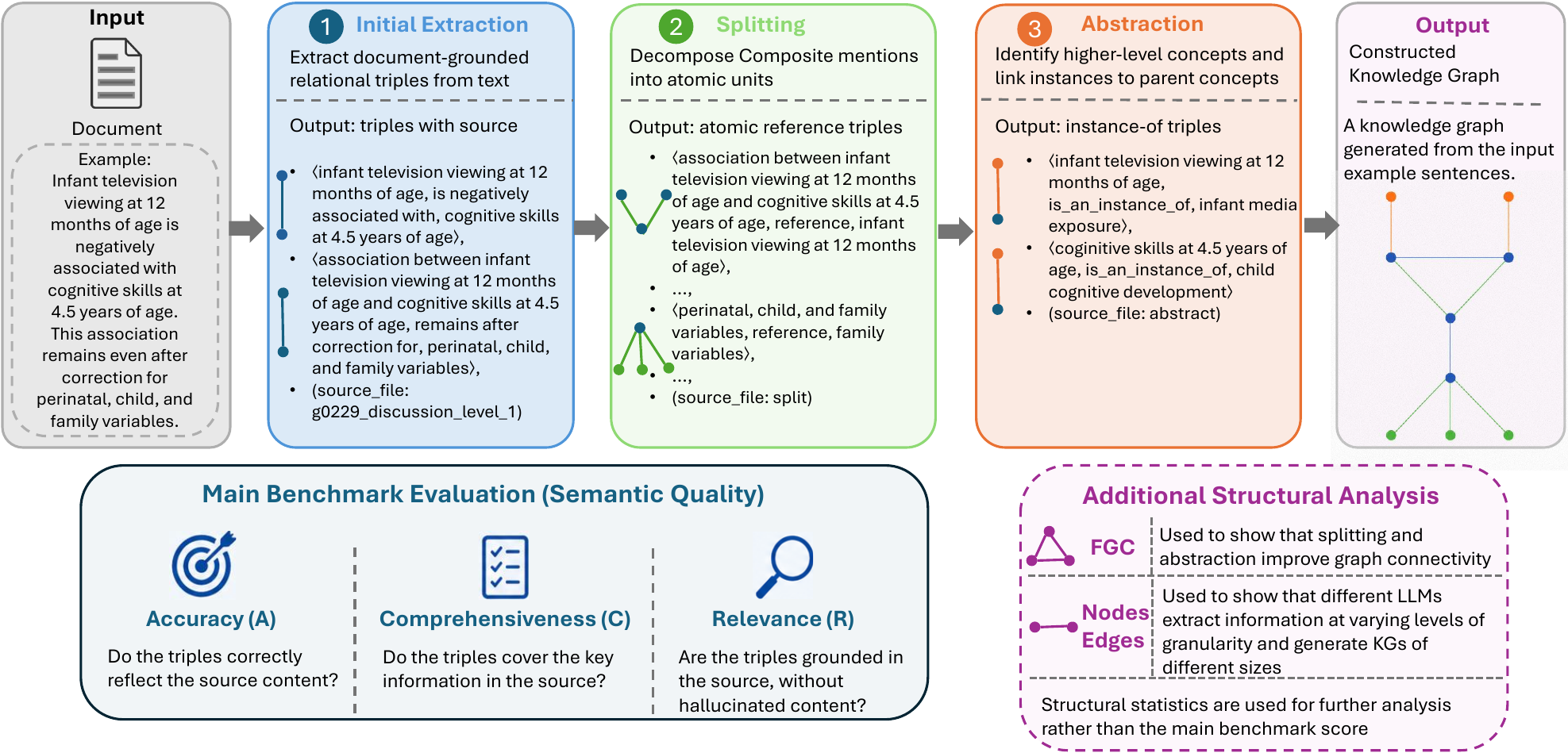}
    \caption{Overview of our hierarchical framework for knowledge graph construction, D2A-HKG. The process involves three steps: (1) extracting information, (2) decomposing composite mentions, and (3) identifying underlying concepts. "source\_file" records the provenance of each triple. For triples produced during the initial extraction stage, it specifies the source document from which the triple was extracted. For triples produced during the decomposition or abstraction stages, "source\_file" is recorded as split or abstract, respectively. In this example, (source\_file: g0229\_discussion\_level\_1) indicates that the triple was extracted from the discussion section of \citet{aishworiya2019television} (paper ID: g0229) during the initial extraction stage. FGC denotes the Fraction in the Giant Component.}
    \label{overview}
\end{figure*}
Knowledge graphs (KGs), a widely used knowledge representation approach comprising entities and relationships, are instrumental in storing factual information \cite{vrandevcic2014wikidata, 10387715}. Leveraging knowledge graphs as a source of external information has shown promising potential in mitigating hallucinations in large language models (LLMs) \cite{agrawal-etal-2024-knowledge, sui2025can} and enabling more effective knowledge editing during LLM usage \cite{chen-etal-2024-llm}. 
At the same time, recent LLM-based knowledge graph construction systems suggest that LLMs can substantially reduce the engineering burden of extraction and normalization \cite{zhang-soh-2024-extract}. Yet one basic question remains insufficiently answered: how well can current frontier LLMs construct document-grounded knowledge graphs in a zero-shot, schema-free setting, where the model must recover useful structure without a predefined ontology? Existing papers provide promising systems, but they do not yet give the field a clean, benchmark-style answer to that question.

In this paper, we propose the Detail-to-Abstract Hierarchical Knowledge Graph (D2A-HKG) framework, illustrated in Figure~\ref{overview}, to benchmark and analyze LLMs' performance across different stages of knowledge graph construction. D2A-HKG decomposes open KG construction into three stages—Initial Extraction, Splitting, and Abstraction—enabling a fine-grained analysis of how well current LLMs perform at each stage.

We apply this framework to CMW-Lit, a collection of document-grounded materials from the medical domain of children’s mental well-being, and benchmark a set of frontier LLMs in a zero-shot setting. This domain is well suited to document-level knowledge graph construction because its evidence is heterogeneous, longitudinal, and distributed across studies. The research papers in CMW-Lit examine interconnected biological, psychological, behavioral, familial, and socioeconomic factors, often within the same longitudinal cohort and across different developmental stages. Their findings also involve complex relationships, including mediation, moderation, temporal associations, and statistically qualified outcomes. These characteristics make CMW-Lit a challenging and practically relevant test bed for evaluating whether knowledge graph construction methods can integrate cross-document evidence while preserving important contextual, quantitative, and provenance information.

A benchmark in this setting requires evaluation that goes beyond exact-match or surface-form overlap. Because schema-free KG construction admits multiple reasonable graph realizations, token-level agreement is often too brittle to capture whether a model extracted the right facts in a useful form \cite{castro-ferreira-etal-2020-2020, zhang-soh-2024-extract}. We therefore make semantic quality the primary benchmark target and evaluate outputs along three dimensions: accuracy, comprehensiveness, and relevance within three knowledge graph construction stages: initial extraction, splitting and abstraction. This choice is consistent with broader evaluation practice in generative NLP \cite{fabbri-etal-2021-summeval} and LLM assessment \cite{ye2024flask}. 
Beyond semantic quality, we also report structural statistics, including graph connectivity and node and edge counts, to support a more comprehensive analysis. These statistics help characterize how different construction stages influence information integration and how unconstrained LLMs can produce substantially different graph structures under the same setting. Our key contributions are as follows:

\begin{enumerate}

\item We introduce D2A-HKG, a benchmark and analysis framework for zero-shot, document-grounded, schema-free knowledge graph construction with LLMs. By decomposing the task into Initial Extraction, Splitting, and Abstraction, D2A-HKG enables fine-grained, stage-wise evaluation of open KG construction. To the best of our knowledge, this is the first work to systematically benchmark LLM-based knowledge graph construction at such a fine-grained, stage-wise level, enabling a detailed analysis of model behavior and performance throughout the construction process.

\item We release CMW-Lit, a challenging dataset for studying KG construction in a high-stakes domain. The generated knowledge graphs can serve as a strong starting point for constructing expert-validated and task-specific knowledge graphs for downstream applications.

\item We benchmark frontier LLMs using semantic quality (accuracy, comprehensiveness, relevance) as the primary evaluation criteria, complemented by structural analysis. We further analyze model-specific patterns in KG extraction under the same prompting setting, highlighting differences in how LLMs construct knowledge graphs and providing empirical observations and resources for future research on LLM-based knowledge graph construction.
    
\end{enumerate}

\section{Related Work}

Knowledge graph construction (KGC) is a composite task that typically involves multiple subtasks, such as information extraction, coreference resolution, entity linking, and knowledge graph completion \cite{ye-etal-2022-generative, 10387715}. With the emergence of LLMs, it has become increasingly feasible to integrate these subtasks within a unified framework, thereby simplifying the development of effective knowledge graphs.

Table \ref{related_works} summarizes several LLM-based KGC frameworks\footnote{See Appendix~\ref{rlc} for the full table, which is omitted from the main text due to space constraints.}. We compare these frameworks in terms of schema requirements, use of few-shot examples, fine-tuning requirements, target domains, and whether they operate at the sentence or document level. Here, the term schema refers not only to a predefined domain schema, but also to predefined extraction mechanisms or structural constraints used during KGC.
As shown in Table \ref{related_works}, existing frameworks adopt different settings. Some rely on predefined schemas, provide examples during construction, target specific domains, require fine-tuning, or support only sentence-level extraction. In addition, their task scopes also vary. For example, EDC \cite{zhang-soh-2024-extract} mainly focuses on relational triplet extraction, whereas CoDe-KG \cite{anuyah-etal-2025-automated} additionally incorporates coreference resolution.

Overall, most existing frameworks rely on relatively complex pipelines or task-specific designs for knowledge graph construction. However, it remains unclear how well state-of-the-art (SOTA) LLMs can perform KGC using simple prompts, particularly in high-stakes domains where accuracy, completeness, and reliability are critical. 

\section{Detail-to-Abstract Hierarchical Knowledge Graph Construction Framework} \label{d2a}
In this paper, we propose a Detail-to-Abstract Hierarchical Knowledge Graph (D2A-HKG) construction framework for evaluating the ability of SOTA LLMs to construct knowledge graphs. Specifically, we conduct KGC in a zero-shot, schema-free, and bottom-up setting. In this setting, no predefined schema or ontology is provided. Instead, relational triples are first extracted from documents, and abstract concepts are then induced from the extracted instances \cite{li2020research, tamavsauskaite2023defining}, as illustrated in Figure \ref{overview}.

To examine LLMs' intrinsic capability for knowledge graph construction, we avoid using complex frameworks. Instead, we rely on simple prompts with clear instructions for information extraction and coreference resolution, which helps reduce ambiguous or incomplete triples. We also provide the initial extraction outputs as references during the subsequent splitting stage, encouraging LLMs to maintain consistent entity and relation representations where possible and thereby improve graph connectivity.

\subsection{Hierarchical Information Extraction} \label{hierarchical_ie}
In a knowledge graph, a node does not necessarily represent a single entity; it may also correspond to a phrase or sentence \cite{wu-etal-2024-coke}. Given the complexity of CMW-Lit, our approach differs from most methods summarized in Table~\ref{related_works}, which typically begin with entity extraction \cite{chen-etal-2024-sac, mo2025kggen}. Instead, we first extract relational triples directly from the document to better preserve sentence-level semantics and contextual information. We refer to this stage as \emph{initial extraction}. Figure~\ref{overview} illustrates an example from \citet{aishworiya2019television}, in which two coarse-grained triples are extracted during this stage. In addition to capturing document meaning, we record source information for each triple, which facilitates verification, traceability, and downstream applications.




Integrating information across multiple sentences at this coarse level is challenging and often produces isolated subgraphs. Therefore, we introduce two fine-grained extraction steps. The first, splitting, decomposes compound node content; for example, "perinatal, child, and family variables" is split into "perinatal variable," "child variable," and "family variable." The second, abstraction, identifies generalized parent concepts; for example, it derives the ancestor concept "infant media exposure" from a specific node such as "infant television viewing at 12 months of age."  Splitting should be performed on all unique entity representations from initial extraction, whereas abstraction should be applied to atomic entities that were not split, as well as to the new entities generated during the splitting stage. Together, these operations support a bottom-up process for constructing a lightweight ontology of concepts, facilitating the integration of information across diverse sources and clarifying the scope of the extracted documents.


\subsection{Coreference-Aware Prompting}
During the initial extraction, our goal is not only to preserve the full meaning of each sentence but also to resolve coreferences that may obscure the underlying semantics. To achieve this, we incorporate explicit instructions in the prompt for the LLMs to perform coreference resolution using the surrounding context. This step is especially crucial when dealing with pronouns or ambiguous phrases, which often refer to complex concepts introduced earlier in the text. By resolving such references, we ensure that the extracted information is both complete and self-contained.

Consider the sentence containing the phrase "this association" shown in Figure \ref{overview}, without coreference resolution, the meaning of this phrase would be unclear. However, through our process, "this association" is resolved and extracted as "association between infant television viewing at 12 months of age and cognitive skills at 4.5 years of age." This allows the resulting knowledge graph to more accurately reflect the intended meaning of the source text and enhances both human interpretability and machine usability in downstream tasks.
\subsection{Information Consistency Prompting}
As observed from the triples extracted during the initial stage, certain semantic relationships are captured, but not fully reflected in the graph structure. For example, the node labeled "association between infant television viewing at 12 months of age and cognitive skills at 4.5 years of age" in Figure \ref{overview} clearly refers to the relationship between two entities: "infant television viewing at 12 months of age" and "cognitive skills at 4.5 years of age." However, despite this implicit connection, these entities are not linked in the knowledge graph at this stage. This disconnect can limit the graph's usefulness for both human interpretation and automated reasoning \cite{DBLP:conf/aaai/MalaviyaBBC20, NGUYEN202056, info13080396, 10.1007/s10462-023-10465-9}.

To address this issue during the splitting stage, we input the triples extracted in the initial stage into the prompt and instruct the LLMs to normalize entity representations—encouraging that identical instances are expressed in the same way across different triples. This helps establish explicit connections between semantically related nodes and improves the overall coherence and connectivity.

\section{Experiments}
In this section, we describe the benchmark datasets, evaluated LLMs, implementation details, and provide an in-depth analysis of the results.
\subsection{Datasets and LLMs}
To benchmark LLM performance in knowledge graph construction in a high-stakes domain, we assembled CMW-Lit, a test set comprising approximately 3,198 sentences from 17 peer-reviewed articles on child mental well-being. As discussed in the introduction, CMW-Lit provides a challenging setting due to its heterogeneous, longitudinal, and context-rich evidence. In terms of scale, it is comparable to the recently proposed MINE benchmark \cite{mo2025kggen}, which contains approximately 2,700 sentences, while presenting greater domain and linguistic complexity.

To evaluate the effectiveness of our pipeline, we further compare our prompting strategy with KGGen \cite{mo2025kggen} using the tasks, datasets (Appendix~\ref{additional_datasets}), and evaluation metrics introduced in its original evaluation. This comparison examines whether our prompt-based approach, described in Section~\ref{d2a}, can achieve competitive performance despite its relatively simple design (Appendix \ref{effectiveness_d2a_hkg}).

To assess the current state of KGC capabilities in SOTA LLMs, we evaluate seven models from four major model families, selected for their strong performance and widespread use: OpenAI's GPT-4o and GPT-5.4\footnote{\url{https://platform.openai.com/docs/models}}; Google's Gemini-3.1-Flash and Gemini-3.1-Pro\footnote{\url{https://ai.google.dev/gemini-api/docs/models}}; Anthropic's Claude-Sonnet-4.6 and Claude-Opus-4.6\footnote{\url{https://claude.com/product/overview}}; and Meta's Llama-4-Maverick\footnote{[\url{https://huggingface.co/meta-llama/Llama-4-Maverick-17B-128E-Instruct}}. We provide no task-specific examples or predefined schemas in the prompts. This zero-shot setting allows us to systematically compare each model's intrinsic ability to extract triples directly from text, with respect to accuracy, comprehensiveness, and relevance.

\subsection{Implementation Details}
During the initial information extraction, we process the text in batches of three sentences from each paper. This helps prevent information loss caused by long input texts \cite{liu-etal-2024-lost, 10.5555/3455716.3455856, edge2024local} and reduces the likelihood of LLMs getting stuck in extended reasoning. For each extraction, we include the previous three sentences as context to improve coherence.

During the splitting stage, we provide the LLMs with the triples extracted in the initial stage. To reduce the cost associated with long prompts, we introduce a lightweight filtering step: the LLM first decides whether further splitting is necessary and only proceeds with splitting when needed.
We perform abstraction for all entities, including those extracted initially and those generated during the splitting stage. Similar to the splitting process, we apply a filtering step to determine whether abstraction is required before executing it. 
\subsection{Evaluation}
Knowledge graphs can be evaluated from both semantic and structural perspectives. Structurally, an effective knowledge graph should have well-connected nodes, minimizing isolated subgraphs. Such connectivity facilitates complete graph traversal and supports effective multi-hop reasoning. To measure structural connectivity quantitatively, we primarily focus on the fraction in giant component, defined as follows:
\begin{equation}
\text{Fraction in Giant Component}
\;=\;
\frac{\lvert C_{\max}\rvert}{\lvert V\rvert}
\end{equation}
where $C_{\max}$ denotes the largest weakly connected component, and $V$ represents the set of all nodes in the graph. Typically, a higher value for this metric is desirable, as it indicates that a larger proportion of nodes belong to the primary connected subgraph and that the graph is less fragmented into small isolated components. However, an excessively connected knowledge graph may also reflect noisy or spurious links. Therefore, careful manual inspection is needed to ensure that improved connectivity does not come at the cost of graph quality.
 
Since our knowledge graph construction is not constrained by predefined schemas or ontologies, and there is no single "correct" graph, we do not provide golden triples for semantic evaluation. Instead, we employ LLMs as evaluators \cite{chen-etal-2024-sac, li2024llmsasjudgescomprehensivesurveyllmbased} to assess multiple quality aspects of the generated knowledge graph. Specifically, we use GPT-4.1\footnote{\url{https://openai.com/index/gpt-4-1/}} as the LLM evaluator. GPT-4.1 is a general-purpose model with strong instruction-following capabilities and broad domain knowledge, making it suitable for applying detailed evaluation rubrics. We evaluate outputs across three stages—initial extraction, splitting, and abstraction—using the metrics of accuracy, comprehensiveness, and relevance. We use a 0–5 Likert-style scale \cite{likert1932technique, fabbri-etal-2021-summeval, qin-etal-2025-r} to capture graded differences in output quality. This scale is widely used in LLM evaluation, including in research on role-playing capabilities. It is appropriate for open-ended KGC evaluation because generated graphs may be partially correct, incomplete, or expressed at different levels of granularity. The detailed evaluation guidelines are provided in Appendix~\ref{evaluation_details}.

\paragraph{Human Validation.}
To select an appropriate LLM judge for evaluating the extracted knowledge graphs, we invited five domain experts to independently assess 47 text chunks and their corresponding 249 triples from the initial extraction stage using the same evaluation guidelines provided to the LLM judges. Among the evaluated judges (GPT-4.1, GPT-5.4, Claude-Opus-4.6, and Gemini-3.1-Pro), GPT-4.1 showed the strongest agreement with the human experts, achieving an exact agreement of 81\% and a Gwet's AC1 \cite{gwet2001handbook} of 0.76, indicating substantial agreement.

We further evaluated the suitability of GPT-4.1 for the splitting and abstraction stages. Five domain experts independently validated GPT-4.1's judgments on 100 samples from each stage. For each stage, the evaluation set comprised 50 positive and 50 negative samples.
Human experts agreed with GPT-4.1's evaluations on 80\% of the splitting samples and 84\% of the abstraction samples. Disagreements often arose because GPT-4.1 applied the evaluation criteria more strictly than the human experts, particularly for negative abstraction cases in which the extractor chose not to abstract an entity. Overall, these results indicate that GPT-4.1 provides reasonably reliable judgments across all three evaluation stages.

To further improve evaluation consistency in the splitting and abstraction stages, we additionally calibrated scores for repeated source--target entity pairs. Because identical source and target entities can be produced independently by different LLM extractors, GPT-4.1 occasionally assigned different score tuples to the same pair across separate evaluations. For each repeated source--target pair, we therefore replaced the individual evaluations with the score tuple most frequently assigned by GPT-4.1 across occurrences. In the event of a tie between score tuples, we selected the more conservative tuple with the lower scores. This calibration reduces run-to-run variation in the LLM judge while preserving a conservative evaluation criterion.

\begin{table*}[t]
\centering
  \renewcommand\arraystretch{1.2}
  \setlength{\tabcolsep}{3.5mm}{
\resizebox{\textwidth}{!}{
\begin{tabular}{@{}lccccccccccc@{}}
\toprule
\multirow{2}{*}{Models} & \multicolumn{3}{c}{Initial Extraction} & \multicolumn{4}{c}{Splitting}    & \multicolumn{4}{c}{Abstraction}   \\
                        & A           & C           & R          & A    & C    & R    & Pos/Neg     & A    & C    & R    & Pos/Neg      \\ \midrule
GPT-4o                  & 4.41        & 4.62        & 4.97       & 4.84 & 4.84 & 4.97 & 1,880/8,427 & 3.33 & 3.29 & 4.86 & 440/12,037   \\
GPT-5.4                 & 4.83        & 4.95        & 4.96       & 4.89 & 4.88 & 4.98 & 756/9,551   & 3.12 & 3.07 & 4.83 & 135/10,973   \\ \midrule
Gemini-3.1-Flash-Lite   & 4.45        & 4.63        & 4.96       & 4.91 & 4.90 & 4.98 & 1,118/9,189 & 3.21 & 3.17 & 4.84 & 1,305/10,277 \\
Gemini-3.1-Pro          & 4.64        & 4.88        & 4.97       & 4.83 & 4.83 & 4.96 & 2,038/8,269 & 3.52 & 3.38 & 4.79 & 4,506/7,370  \\ \midrule
Claude-Sonnet-4.6       & 4.57        & 4.88        & 4.94       & 4.93 & 4.92 & 4.99 & 573/9,734   & 3.42 & 3.39 & 4.86 & 2,490/8,429  \\
Claude-Opus-4.6         & 4.59        & 4.83        & 4.97       & 4.92 & 4.92 & 4.99 & 585/9,722   & 3.24 & 3.18 & 4.84 & 1,164/9,631  \\ \midrule
Llama-4-Maverick        & 3.98        & 4.28        & 4.91       & 4.64 & 4.63 & 4.90 & 3,683/6,624 & 3.79 & 3.70 & 4.78 & 7,570/5,524  \\ \bottomrule
\end{tabular}%
}}
\caption{Performance of the hierarchical information extraction pipeline, including initial extraction, splitting, and abstraction. Initial extraction is evaluated using accuracy (A), comprehensiveness (C), and relevance (R). For splitting and abstraction, performance is computed over all entities, regardless of whether they are split or abstracted. Pos/Neg denotes the number of positive entities that underwent splitting/abstraction and negative entities that did not undergo splitting/abstraction, respectively.
}
\label{semantic_main}
\end{table*}

\subsection{Semantic Analysis}
Table~\ref{semantic_main} presents the accuracy, comprehensiveness, and relevance scores across the initial extraction, splitting, and abstraction stages. Because different LLMs generate different numbers of triples during initial extraction, we use the entities extracted by GPT-5.4 as a common input to the splitting stage to enable a relatively fair comparison across models. For the abstraction stage, however, we evaluate each model on the entity set resulting from its own splitting decisions, consisting of the original entities that were not split and the new entities produced through splitting. Using an identical entity set for abstraction would decouple abstraction from the preceding splitting stage and would not reflect the actual sequential behavior of the pipeline.

\paragraph{LLMs show strong zero-shot potential for knowledge graph construction.}
As shown in Table~\ref{semantic_main}, most evaluated models achieve average scores of approximately 4 across the evaluated dimensions, indicating generally high-quality outputs with only minor errors. In addition, most models effectively exclude information that is unrelated to the research focus of CMW-Lit, such as author affiliations, funding statements, and other document metadata. Hallucinated or clearly irrelevant content is also rare, resulting in near-perfect relevance scores across multiple stages. These findings demonstrate the strong potential of recent LLMs for zero-shot, document-level knowledge graph construction. Nevertheless, some models still exhibit difficulties in strictly following the required output format. For example, LLama-4-Maverick occasionally produces malformed JSON outputs, preventing some extracted triples from being parsed and stored correctly and thereby reducing its evaluation scores. This suggests that robust parsing mechanisms, constrained generation, or post-processing may still be necessary for reliable deployment.

\paragraph{LLMs exhibit diverse behaviors in the initial extraction stage.}
As shown in Table~\ref{semantic_main} and Table~\ref{structure_main}, GPT-5.4 achieves the strongest overall performance in the initial extraction stage, obtaining the highest accuracy, comprehensiveness, and relevance scores. It also generates the largest number of triples per text chunk, indicating that its broader extraction coverage does not substantially compromise accuracy. Gemini-3.1-Pro ranks second, achieving strong coverage and high comprehensiveness, although its accuracy is slightly lower than that of GPT-5.4. Claude-Opus-4.6 and Claude-Sonnet-4.6 exhibit comparable overall performance, with Claude-Sonnet-4.6 achieving slightly higher comprehensiveness and Claude-Opus-4.6 slightly higher accuracy. In contrast, GPT-4o and Gemini-3.1-Flash-Lite exhibit more selective extraction behavior and generate fewer triples per text chunk, which may partly explain their lower comprehensiveness. LLama-4-Maverick shows the weakest overall performance at this stage. Although it does not generate the fewest triples, its lower accuracy and comprehensiveness indicate a higher prevalence of imprecise extractions, omitted information, and malformed structured outputs. Overall, these results suggest that effective initial extraction requires a balance between broad information coverage and faithful representation of the source text, with GPT-5.4 achieving the best balance.

\paragraph{LLMs exhibit substantially different splitting and abstraction strategies.}
The results from the splitting and abstraction stages reveal substantial differences in how aggressively individual LLMs transform entities, suggesting that downstream performance is influenced not only by overall model capability but also by model-specific transformation strategies. Claude-Sonnet-4.6 and Claude-Opus-4.6 adopt relatively conservative splitting strategies, producing fewer split entities but generally achieving high-quality decompositions when splitting is performed. Claude-Sonnet-4.6 is particularly effective in the abstraction stage, performing abstraction more frequently than Claude-Opus-4.6 while maintaining high evaluation scores, indicating a favorable balance between transformation coverage and precision. GPT-5.4 is also conservative, especially in abstraction, where it performs relatively few abstraction operations. While this behavior reduces the risk of introducing inappropriate higher-level concepts, it may also omit potentially useful abstractions.
In contrast, GPT-4o and Gemini-3.1-Pro adopt more aggressive splitting strategies and consequently generate substantially more split entities. However, their lower splitting scores suggest that they sometimes decompose entities that are not sufficiently composite to warrant splitting. Gemini-3.1-Flash-Lite exhibits a more balanced strategy, performing fewer transformations than Gemini-3.1-Pro while maintaining relatively high quality among the entities it chooses to split or abstract. LLama-4-Maverick is the most aggressive model in both stages. Although this behavior increases the number of transformed entities and may enhance graph granularity or connectivity, it also introduces more unnecessary or semantically questionable transformations, resulting in lower average splitting quality.
We further observe that GPT-4.1, used as the LLM judge, tends to favor abstraction and assign higher abstraction-stage scores to models that perform abstraction more frequently. This tendency is consistent with our human-validation findings, suggesting that the judge may exhibit a preference toward valid abstraction operations rather than purely conservative abstraction behavior.

\begin{table*}[t]
  \centering
  \renewcommand\arraystretch{1.2}
  \setlength{\tabcolsep}{5mm}{
\resizebox{\textwidth}{!}{
    \begin{tabular}{lcccccc}
\toprule
\multirow{2}{*}{Models} & \multicolumn{2}{c}{Initial Extraction}   & \multicolumn{2}{c}{With Splitting}  & \multicolumn{2}{c}{With Splitting \& Abstraction}                        \\ 
                        & \multicolumn{1}{c}{$\text F_{GC}$} & Nodes/Edges & \multicolumn{1}{c}{$\text F_{GC}$} & Nodes/Edges & \multicolumn{1}{c}{$\text F_{GC}$} & Nodes/Edges \\ \midrule
GPT-4o           & \multicolumn{1}{c}{0.006}   & 6,890/3,994  & \multicolumn{1}{c}{0.157}                    & 10,236/8,117  & \multicolumn{1}{c}{0.207}                    &  11,164/9,204  \\
GPT-5.4                  & \multicolumn{1}{c}{0.004}   & 10,307/6,601  & \multicolumn{1}{c}{0.018}                    & 11,864/8,812 & \multicolumn{1}{c}{0.020}                   &  11,920/8,947  \\
\midrule
Gemini-3.1-Flash-Lite        & \multicolumn{1}{c}{0.013}   & 7,002/4,308  & \multicolumn{1}{c}{0.014}                    &  8,673/6,355 & \multicolumn{1}{c}{0.063}                   &  9,301/7,321 \\
Gemini-3.1-Pro        & \multicolumn{1}{c}{0.159}   &  8,393/6,346   & \multicolumn{1}{c}{0.438}                    & 10,147/9,548 & \multicolumn{1}{c}{0.732}                   & 12,486/13,789    \\ \midrule
Claude-Sonnet-4.6        & \multicolumn{1}{c}{0.023}   &  9,684/6,347 & \multicolumn{1}{c}{0.050}                    & 10,662/7,865  & \multicolumn{1}{c}{0.075}                   &  13,554/11,302  \\
Claude-Opus-4.6        & \multicolumn{1}{c}{0.149}   & 8,423/5,954  & \multicolumn{1}{c}{0.218}                    &  9,166/7,067 & \multicolumn{1}{c}{0.326}                   &  9,570/7,879   \\ \midrule
Llama-4-Maverick          & \multicolumn{1}{c}{0.137}   &  7,364/5,138 & \multicolumn{1}{c}{0.611}                    & 10,241/10,779   & \multicolumn{1}{c}{0.665}                    & 16,005/17,044   \\ \bottomrule
\end{tabular}}}%
\caption{Structural statistics of all LLM-generated knowledge graphs, including the number of nodes and edges, as well as the fraction in the giant component ($\text F_{GC}$).}
  \label{structure_main}%
\end{table*}%

\subsection{Structural Analysis}

Table~\ref{structure_main} reports structural statistics of the knowledge graphs generated from a subset of the CMW-Lit dataset, including the numbers of nodes and edges and the proportion of nodes contained in the giant component. It shows that under the same zero-shot, schema-free setting, different LLMs produce substantially different knowledge graphs from the same input. In addition, incorporating split and abstract triples generally improves graph connectivity, as indicated by the increased fraction of nodes in the giant weakly connected component. 

Consistent with our goal of hierarchical information extraction, these results suggest that the second-stage triples help connect previously fragmented parts of the graph. Specifically, split triples connect compound entities with their component entities, while abstract triples link specific entities to broader conceptual categories. Together, they serve as structural connectors by introducing intermediate links that allow separate subgraphs to be merged into larger connected components. This indicates that split and abstract triples can facilitate information integration across different sources.
However, higher connectivity is not always inherently better and connectivity should be improved in a controlled manner.

    


\section{Future Directions}

Although LLMs demonstrate strong potential for zero-shot, schema-free knowledge graph construction, our analysis also highlights several directions that require further investigation.

\paragraph{Selective extraction of salient information.}
Future work should investigate how to guide LLMs to selectively identify and extract the most salient information, rather than processing all sentences indiscriminately. For knowledge graph construction from research papers, models should prioritize information central to the scientific contribution, such as the main research questions, methodological settings, key findings, while filtering out peripheral or background statements that contribute little to the core knowledge representation. For example, a sentence such as "Although infant media exposure has received attention for its implications for child development, upstream risk factors contributing to media exposure have rarely been explored" may provide useful motivation for the study but is less critical for inclusion in a knowledge graph focused on the study's substantive findings.
\paragraph{Improved graph connectivity and consistency.}
Although the extracted triples are often semantically correct and clearly expressed, information originating from the same section is not always well connected in the resulting knowledge graph. Semantically related findings may therefore be represented as separate local structures without explicit links between them. Splitting and abstraction can partially improve connectivity by introducing finer-grained entities and higher-level concepts, but related information may still require traversing multiple hops to be identified or retrieved. Future work of initial extraction should therefore focus not only on accurate information extraction, but also on maintaining coherent cross-triple relationships, consistent entity representations, and stronger links among related findings throughout the graph.

\paragraph{More reliable splitting and abstraction.}
The splitting and abstraction stages require further investigation, as LLMs may struggle to determine the appropriate degree of transformation and can therefore split or abstract entities either too aggressively or too conservatively. Accurate splitting can decompose complex entities into more precise atomic concepts while preserving their original semantics, whereas effective abstraction can organize related entities into coherent higher-level categories without introducing overly broad or semantically inappropriate concepts. Together, these operations can improve the granularity, organization, and interpretability of the resulting graph, while also facilitating knowledge integration and alignment with external knowledge sources.

\paragraph{Scalable long-context knowledge graph construction.}
How to input long documents into LLMs while obtaining accurate, comprehensive, and relevant results remains an open challenge. Processing only a few sentences at a time may miss important cross-sentence and cross-paragraph dependencies, especially when pronouns or referring expressions point to entities outside the local context. Providing LLMs with all existing entities in a knowledge graph may help unify entity representations, but this strategy becomes infeasible for large-scale graphs. Therefore, effective filtering and retrieval strategies are needed to identify the most relevant existing entities. At the same time, LLMs must distinguish between entities that appear similar but should not be merged, avoiding inappropriate entity merging.

\section{Conclusion}
In this paper, we proposed D2A-HKG, a hierarchical framework for evaluating zero-shot, schema-free knowledge graph construction with SOTA LLMs. We introduced CMW-Lit, a challenging dataset featuring heterogeneous evidence, domain-specific terminology, complex scientific relationships, and information distributed across documents and sections.
We evaluated the resulting knowledge graphs from semantic and structural perspectives. Our results show that recent LLMs have strong potential for scalable knowledge graph construction but exhibit distinct behaviors across extraction, splitting, and abstraction. Domain-expert verification remains important for developing gold-standard graphs that faithfully represent complex scientific knowledge.
Finally, we discussed directions for improving LLM-based knowledge graph construction and believe that our study will support future research on knowledge graph construction and KG-based downstream applications.

\section*{Limitations}
The primary focus of our paper is to evaluate the performance of LLMs in zero-shot, schema-free knowledge graph construction and to identify their strengths and limitations. Accordingly, we do not aim to propose a highly sophisticated knowledge graph construction pipeline; the development of more advanced pipelines is left for future work. In addition, although we validate and calibrate LLM-based evaluation whenever possible, we acknowledge that LLM evaluators are not perfect, particularly in the challenging zero-shot, schema-free setting. Nevertheless, given the scale and open-ended nature of the task, LLM-based evaluation provides the most practical and scalable evaluation approach available to us.

\section*{Ethical Considerations}
LLMs may produce incomplete, oversimplified, or inaccurate triples, especially when handling conditional, non-binary, or multi-layered relationships in scientific texts. Such errors could misrepresent the original findings if used without careful validation. In this work, we use peer-reviewed publications as source documents and evaluate generated triples in terms of accuracy, comprehensiveness, and relevance. We also emphasize that expert verification remains essential, particularly in high-stakes domains.

Finally, knowledge graphs constructed from scientific literature may inherit limitations or biases from the original studies, such as population-specific findings or measurement constraints. Users of CMW-Lit and D2A-HKG should interpret the extracted knowledge within the context of the source papers and avoid using generated graphs for clinical, educational, or policy decisions without appropriate human oversight.

\bibliography{custom}

\appendix
\clearpage
\newpage
\section{Related Work Comparison} \label{rlc}
We review a range of knowledge graph construction studies and summarize them in Table~\ref{related_works}, comparing them across multiple dimensions. As shown in the table, existing frameworks adopt substantially different experimental settings: some rely on predefined schemas, some use few-shot prompting, some require model fine-tuning, while others focus exclusively on sentence-level extraction. These differences preclude a direct comparison between our schema-free, zero-shot D2A-HKG framework and these approaches.

\section {Evaluation Details} \label{evaluation_details}
We evaluate outputs across three stages—initial extraction, splitting, and abstraction—using the metrics of accuracy, comprehensiveness, and relevance. 
For example, in the initial extraction stage, these metrics are defined as follows:
\begin{itemize}
\item \textbf{Accuracy}: Do the triples' head/tail texts and their properties exactly reflect the sentence (correct meaning, correct boundaries, factual alignment), and are relations precisely labeled?
\item \textbf{Comprehensiveness}: Do the triples collectively cover all distinct factual/relational information present in the sentence (no important fact missing)?
\item \textbf{Relevance}: Are the triples free of information not grounded in the sentence (no hallucinated entities, properties, or relations)?
\end{itemize}

The general scoring guidance is provided in Table~\ref{tab:scoring_guidance}.

\begin{table*}[]
\centering
\renewcommand\arraystretch{1.0}
\setlength{\tabcolsep}{1.0mm}{
\resizebox{0.8\textwidth}{!}{
\begin{tabular}{lccccc}   
\toprule
Framework                                                         & Schema & FSL & Domain    & Finetune & Doc/Sent \\
\midrule
\citet{agrawal-etal-2022-large}                 & \checkmark      & \checkmark                 & Clinical  & \ding{55}  & Sentence       \\  

CodeKGC \cite{10.1145/3641850}                  & \checkmark   & \ding{55}     & General   & \checkmark    & Sentence          \\
EDC \cite{zhang-soh-2024-extract}               & \ding{55} & \checkmark   & General   & \checkmark    & Sentence          \\
SAC-KG \cite{chen-etal-2024-sac}                & \ding{55}   & \checkmark      & Rice      & \checkmark    & Document\\          
CoDe-KG \cite{anuyah-etal-2025-automated}       & \ding{55}  & \checkmark & General   & \ding{55}  & Sentence          \\

Tree-KG \cite{niu-etal-2025-tree}               & \checkmark   & \checkmark    & Textbook  & \ding{55}  & Document   \\      

iText2KG   \cite{10.1007/978-981-96-0573-6_16} & \checkmark    & \ding{55}   & General   & \ding{55}     & Document         \\
KGGen \cite{mo2025kggen}                        & \ding{55}   & \ding{55}    & General   & \ding{55} & Document          \\
AutoSchemaKG \cite{bai2025autoschemakg}         & \ding{55}  & \checkmark & General   & \ding{55} & Document          \\
TRACE-KG \cite{abolhasani2026beyond}            & \checkmark   & \checkmark    & Technical & \ding{55}   & Document         \\
Wikontic   \cite{chepurova-etal-2026-wikontic} & \checkmark& \checkmark       & General   & \ding{55}        & Sentence          \\
SocraticKG \cite{choi2026socratickg}            & \checkmark & \ding{55}   & General   & \ding{55}  & Document          \\
\bottomrule
\end{tabular}
}}
\caption{Comparison of knowledge graph construction frameworks in terms of schema requirements, few-shot learning (FSL), fine-tuning, target domains, and document-level applicability.}
\label{related_works}
\end{table*}

\begin{table}[t]
\centering
\resizebox{\columnwidth}{!}{
\begin{tabular}{cl}
\toprule
\textbf{Score} & \textbf{Scoring Guidance} \\
\midrule
0 & Wholly incorrect / none satisfied. \\
1 & Very poor (few fragments correct). \\
2 & Partial but major issues. \\
3 & Moderate quality with noticeable gaps or errors. \\
4 & High quality with only minor issues. \\
5 & Perfect (meets definition fully). \\
\bottomrule
\end{tabular}
}
\caption{Scoring guidelines for all metrics.}
\label{tab:scoring_guidance}
\end{table}

\paragraph{Reproducibility Details}
We evaluate GPT-4o (2024-11-20), GPT-5.4 (2026-03-05), Claude Sonnet 4.6, Claude Opus 4.6, Gemini 3.1 Flash-Lite, Gemini 3.1 Pro Preview, and Llama 4 Maverick 17B 128E Instruct FP8. We set the temperature to 0 whenever supported by the model. 
We performed the LLM-based extraction and evaluation once on the full dataset. However, we repeated the extraction process twice for GPT-5.4 and Claude Opus 4.6 on one paper and evaluated both runs using GPT-4.1. The results showed only minor run-to-run variation: the mean absolute score differences were 0.013 for GPT-5.4 and 0.043 for Claude Opus 4.6, with population variances of 0.00027 and 0.00376, respectively. 

\section{Additional Evaluation Datasets} \label{additional_datasets}
\begin{itemize}
\item \textbf{MINE} \cite{mo2025kggen}. MINE \cite{mo2025kggen} consists of 100 articles spanning diverse domains, including Arts, Culture \& Society, Science, Technology, Psychology/Human Experience, and History \& Civilization.

\item \textbf{WikiQA} \cite{yang-etal-2015-wikiqa}. To evaluate the effectiveness of our approach in downstream tasks, we conduct a retrieval-augmented generation (RAG) evaluation on the WikiQA dataset, following \citet{mo2025kggen}. For a focused and cost-efficient evaluation, we assess answer accuracy on 242 questions associated with 239 Wikipedia articles from the WikiQA test set.
\end{itemize}
\section{Generality and Effectiveness of D2A-HKG} \label{effectiveness_d2a_hkg}

Although our study primarily focuses on evaluating LLMs' zero-shot knowledge graph construction capabilities, we further compare D2A-HKG with KGGen~\cite{mo2025kggen}, a recent knowledge graph construction framework, to demonstrate the generality and effectiveness of our proposed D2A-HKG.
KGGen introduces two evaluation tasks: MINE-1, which assesses knowledge retention, and MINE-2, which evaluates KG-assisted retrieval-augmented generation (RAG), together with a corresponding KG construction approach.

We first evaluate our approach on MINE-1, a dataset introduced by KGGen. MINE-1 consists of 100 articles from diverse domains, including arts, culture, and society. Following KGGen, we adopt its knowledge retention metric for evaluation and compare our approach with KGGen, the classical knowledge graph construction method OpenIE \cite{angeli-etal-2015-leveraging}, and Microsoft's GraphRAG \cite{larson2024graphrag, edge2024local}. To ensure a fair comparison, we reuse the publicly released evaluation code and follow the same evaluation settings, including the same LLM and default retrieval model. Thus, the knowledge graph construction method is the primary varying factor across the compared approaches. As shown in Table~\ref{comparison_with_mine1}, our approach achieves a knowledge retention score of 83\% when using GPT-4o for extraction, substantially outperforming KGGen with GPT-4o (66\%) and KGGen with Claude Sonnet 3.5\footnote{\url{https://www.anthropic.com/news/claude-3-5-sonnet}} (73\%).

To further examine whether the knowledge graphs constructed by our pipeline can effectively support downstream applications, we compare our approach with KGGen on the WikiQA test set, as shown in Table~\ref{comparison_with_mine2}. Because we could not locate the code used for KG-assisted RAG evaluation, we reproduced the evaluation pipeline based on the methodological details provided in the KGGen paper and evaluated both approaches on 239 Wikipedia articles and 242 questions from the WikiQA test set. Gemini 2.5 Flash was used for knowledge graph extraction, while GPT-4o was used for evaluation. Apart from the knowledge graph construction method, all other experimental settings were kept consistent to ensure a fair comparison. We consider two evaluation settings: the default KGGen setting, in which retrieved triples are provided together with their associated text chunks, and a triple-only setting, in which question answering relies solely on the retrieved triples. Our approach outperforms KGGen under both settings. The improvement is particularly pronounced in the triple-only setting, where our approach achieves approximately 10 percentage points higher QA accuracy, suggesting that the constructed knowledge graphs preserve information more effectively even without access to the original text chunks.

Overall, these results demonstrate that, despite its simplicity, D2A-HKG is competitive with existing knowledge graph construction methods. Moreover, the constructed knowledge graphs not only retain document-level knowledge effectively but also provide useful structured knowledge for downstream question answering.

\begin{table}[t]
\centering
\resizebox{0.9\columnwidth}{!}{
\begin{tabular}{@{}llc@{}}
\toprule
Approach               & Models            & MINE-1 Score  \\ \midrule
OpenIE                 & GPT-4o            & 0.30          \\ \midrule
GraphRAG               & GPT-4o            & 0.48          \\ \midrule
\multirow{3}{*}{KGGen} & Gemini-2.0-Flash  & 0.44          \\
                       & GPT-4o            & 0.66          \\
                       & Claude-Sonnet-3.5 & 0.73          \\ \midrule
D2A-HKG                & GPT-4o            & \textbf{0.83} \\ \bottomrule
\end{tabular}%
}
\caption{Experiment conducted on the MINE-1 dataset, which comprises 100 articles spanning diverse topics.}
\label{comparison_with_mine1}
\end{table}

\begin{table}[]
\centering
\resizebox{0.85\columnwidth}{!}{
\begin{tabular}{@{}llc@{}}
\toprule
Input                                & Approach & Accuracy         \\ \midrule
\multirow{2}{*}{With Text Chunks}    & KGGen    & 82.23\%          \\
                                     & D2A-HKG  & \textbf{83.88\%} \\ \midrule
\multirow{2}{*}{Without Text Chunks} & KGGen    & 60.74\%          \\
                                     & D2A-HKG  & \textbf{70.66\%} \\ \bottomrule
\end{tabular}%
}
\caption{Experiment on the WikiQA test set, which contains 239 Wikipedia articles and 242 questions. Gemini-2.5-Flash is used for extraction, and GPT-4o is used for evaluation.
}
\label{comparison_with_mine2}
\end{table}

\section{Top-Down and Bottom-Up Knowledge Graph Construction Approaches}
Knowledge graphs are effective tools for integrating information from diverse sources, and their construction should facilitate the clear and structured representation of that information. Typically, there are two main approaches to building a knowledge graph: the top-down and bottom-up methods \cite{tamavsauskaite2023defining}. The top-down approach is commonly used in domain-specific knowledge graph construction, where a well-defined ontology exists, and the graph is populated based on that ontology. In contrast, the bottom-up approach is often applied to general-purpose knowledge graphs, where the information is broader in scope and a clear, predefined ontology is lacking. However, with the rise of large language models (LLMs), even domain-specific chatbots now demand a higher level of comprehensiveness. This makes it increasingly difficult to rely on a fixed and limited ontology when constructing knowledge graphs for such applications. As a result, in this paper, we adopt the bottom-up approach to build our knowledge graph, allowing for greater flexibility and adaptability in representing knowledge.

\section{The Evaluation Disagreements between Human Experts and LLM Evaluator}

As described in the human validation study, five domain experts independently evaluated 47 text chunks and their corresponding 249 triples extracted by GPT-5.4 during the initial extraction stage. Table~\ref{human_llm_inconsistency} summarizes the disagreements between the human experts and GPT-4.1. Most disagreements occurred in the accuracy dimension, followed by relevance, whereas relatively few were observed for comprehensiveness.

\begin{table}[]
\resizebox{\columnwidth}{!}{
\begin{tabular}{lcc}
\toprule
Dimension         & \begin{tabular}[c]{@{}l@{}}Disagreements/\\ Pairwise ratings\end{tabular} & \begin{tabular}[c]{@{}l@{}}Share of all\\ disageements\end{tabular} \\ \midrule
Accuracy          & 78/235                                                                    & 58.6\%                                                              \\ \midrule
Comprehensiveness & 21/235                                                                    & 15.8\%                                                              \\ \midrule
Relevance         & 34/235                                                                    & 25.6\%                                                              \\ \bottomrule
\end{tabular}%
}
\caption{Disagreements Between Five Human Experts and GPT-4.1 in the Initial Extraction Stage}
\label{human_llm_inconsistency}
\end{table}
\section{Additional Insights}
In addition to the above analysis, we also found that reasoning-oriented models, such as o4-mini, are capable of integrating information across multiple sentences. For example, it can extract and combine information from two unconnected sentences, as shown below:
\begin{itemize}
    \item Sample Sentences from \citet{fok2016comparison}: Participants were part of a prospective birth cohort study that recruited 1,247 pregnant women (57.2\% Chinese, 25.5\% Malay, and 17.3\% Indian) during their first trimester....Most participants reported that they followed confinement practices during the first 3 weeks postpartum (Chinese: 96.4\%, Malay: 92.4\%, Indian: 85.6\%). 
    \item Head Entity: Chinese pregnant women (57.2\% of the cohort)
    \begin{itemize}
        \item percentage\_of\_cohort: 57.2
        \item adherence\_rate\_first\_3\_weeks: 96.4
    \end{itemize}
    \item Relation: followed
    \item Tail Entity: confinement practices during the first 3 weeks postpartum
\end{itemize}

However, we also observed several shortcomings across the evaluated models. For instance, GPT-3.5-Turbo occasionally fails to adhere to the specified output template, generates repetitive triples, and sometimes struggles to distinguish the target text from its preceding context. Llama-3.1-8B\footnote{\url{https://ai.meta.com/blog/meta-llama-3-1/}} often has difficulty following the output format and occasionally produces vague or uninformative relations, such as "in." Similarly, Llama-3.3-70B-Instruct\footnote{\url{https://www.llama.com/docs/model-cards-and-prompt-formats/llama3_3/}} exhibits issues with template compliance and tends to overthink whether the text requires further splitting or abstraction.

\begin{figure*}[t!]
    \centering
    \includegraphics[width=2\columnwidth]{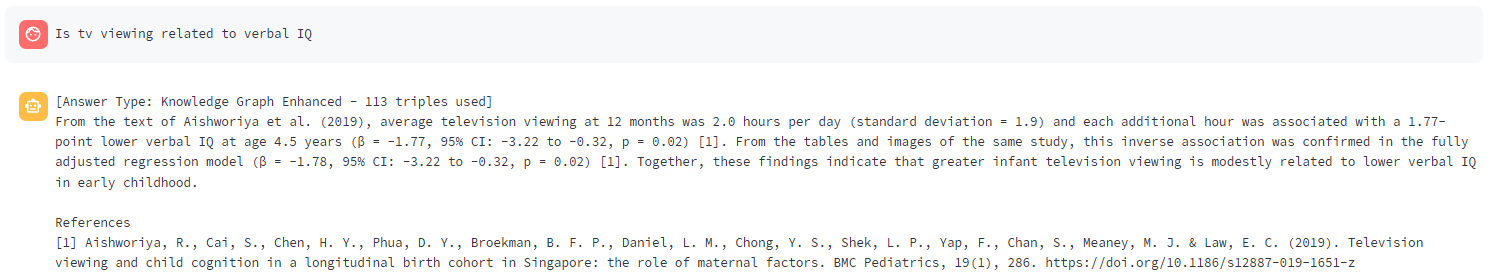}
    \caption{Example 1 of Knowledge Graph–Supported Question Answering.}
    \label{qa_sample_1}
\end{figure*}

\begin{figure*}[t!]
    \centering
    \includegraphics[width=2\columnwidth]{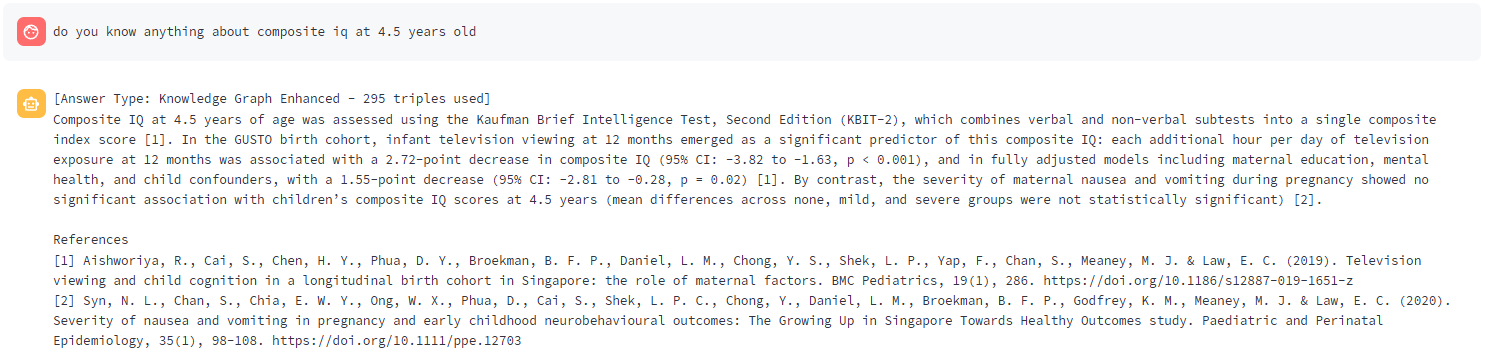}
    \caption{Example 2 of Knowledge Graph–Supported Question Answering.}
    \label{qa_sample_2}
\end{figure*}

\section{Knowledge Graph Visualization}
\begin{figure*}[t!]
    \centering
    \begin{subfigure}[t]{0.24\textwidth}
        \centering
        \includegraphics[width=\textwidth]{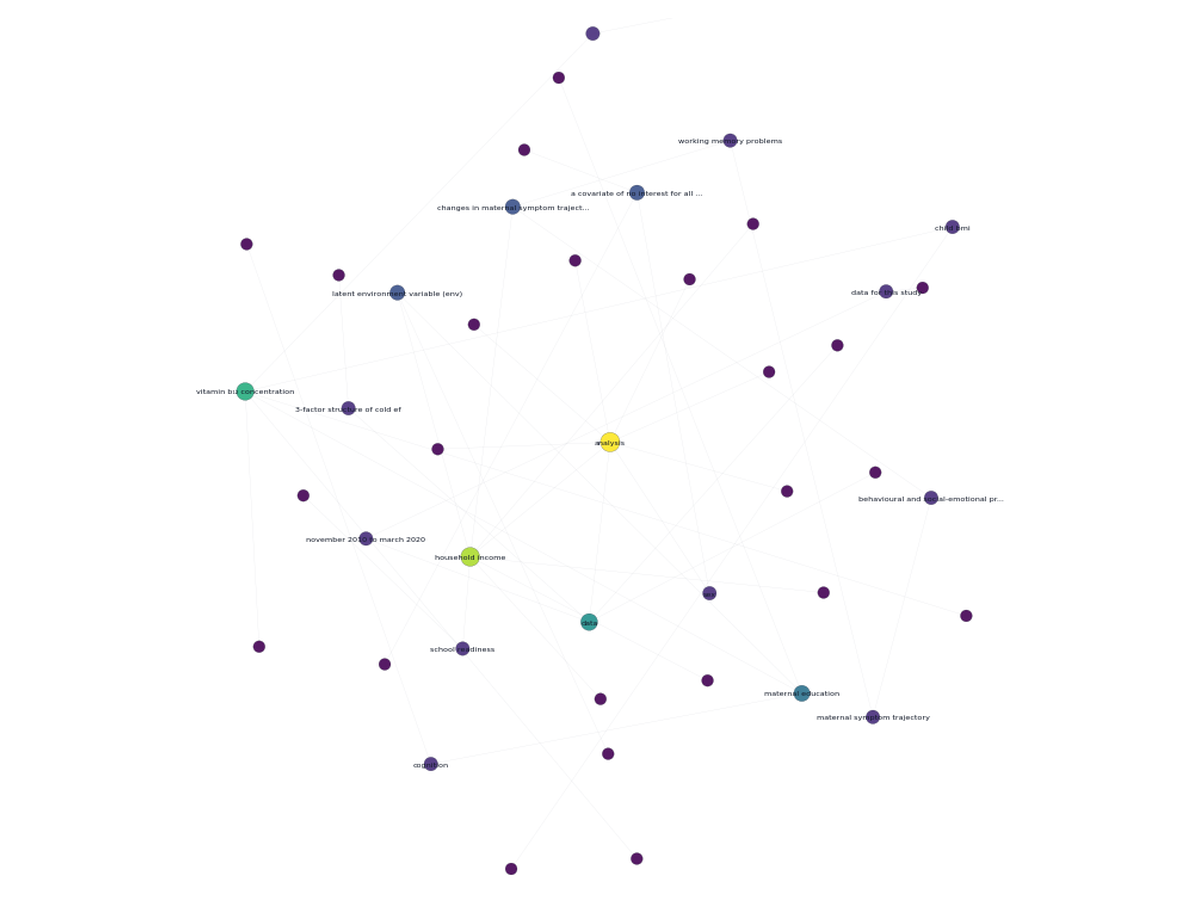}
        \caption{GPT-4o initial extraction.}
        \label{fig:gpt4o-initial}
    \end{subfigure}
    \hfill
    \begin{subfigure}[t]{0.24\textwidth}
        \centering
        \includegraphics[width=\textwidth]{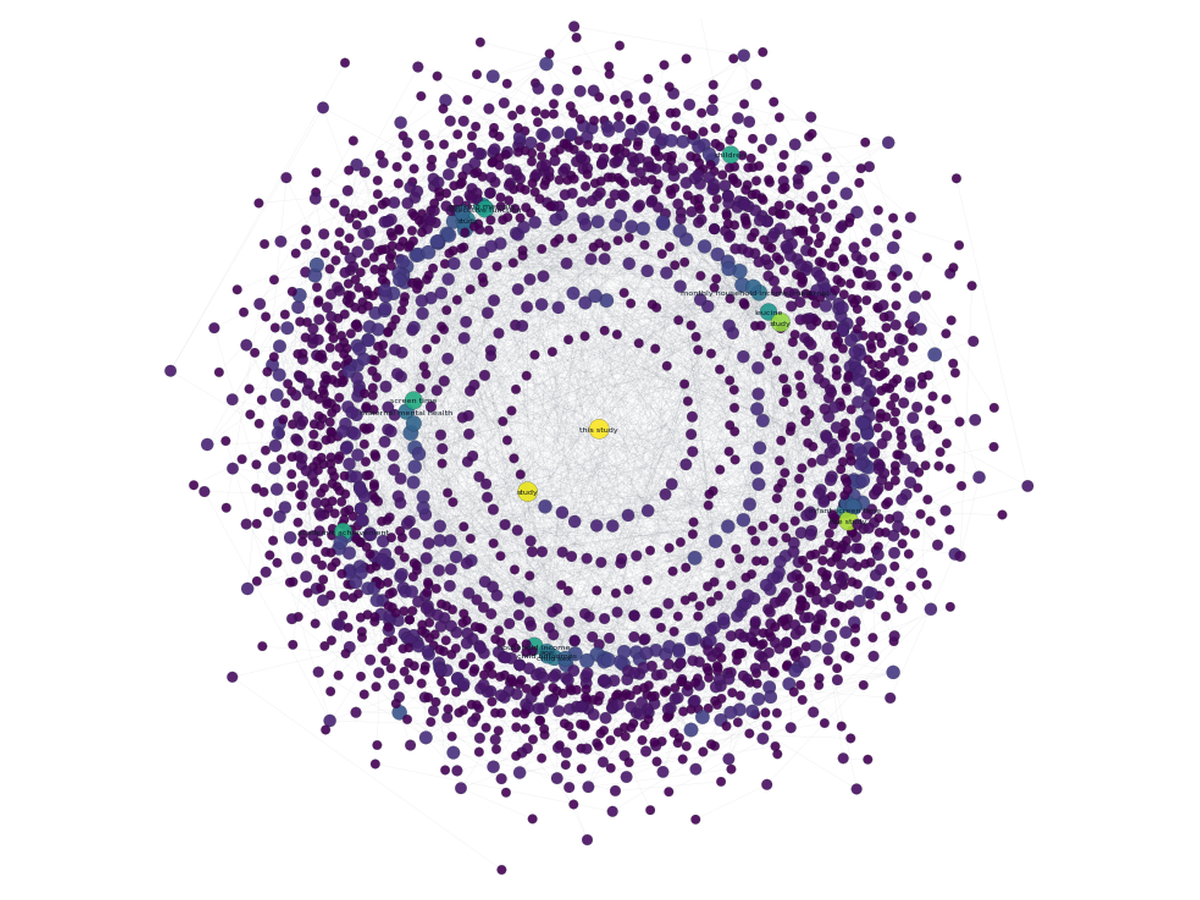}
        \caption{GPT-4o full extraction.}
        \label{fig:gpt4o-full}
    \end{subfigure}
    \hfill
    \begin{subfigure}[t]{0.24\textwidth}
        \centering
        \includegraphics[width=\textwidth]{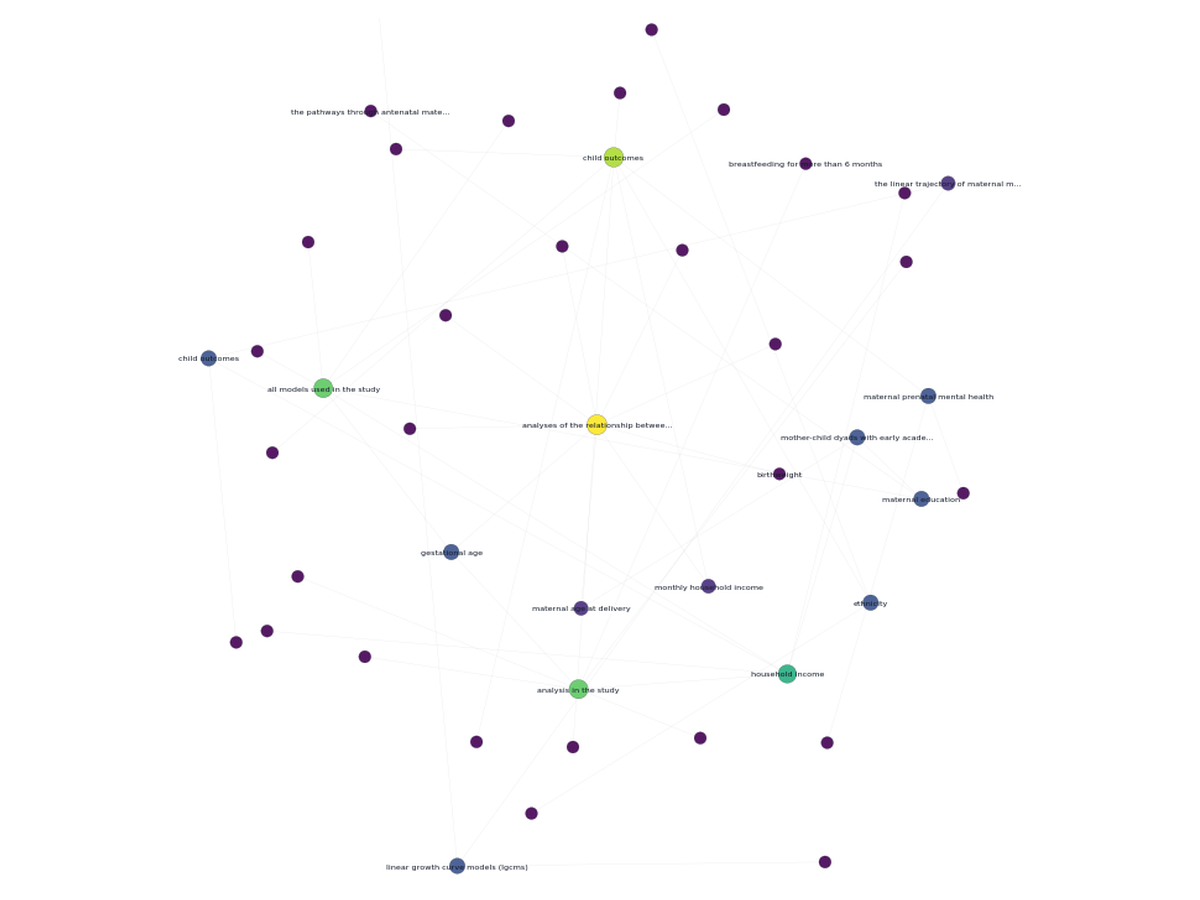}
        \caption{GPT-5.4 initial extraction.}
        \label{fig:gpt54-initial}
    \end{subfigure}
    \hfill
    \begin{subfigure}[t]{0.24\textwidth}
        \centering
        \includegraphics[width=\textwidth]{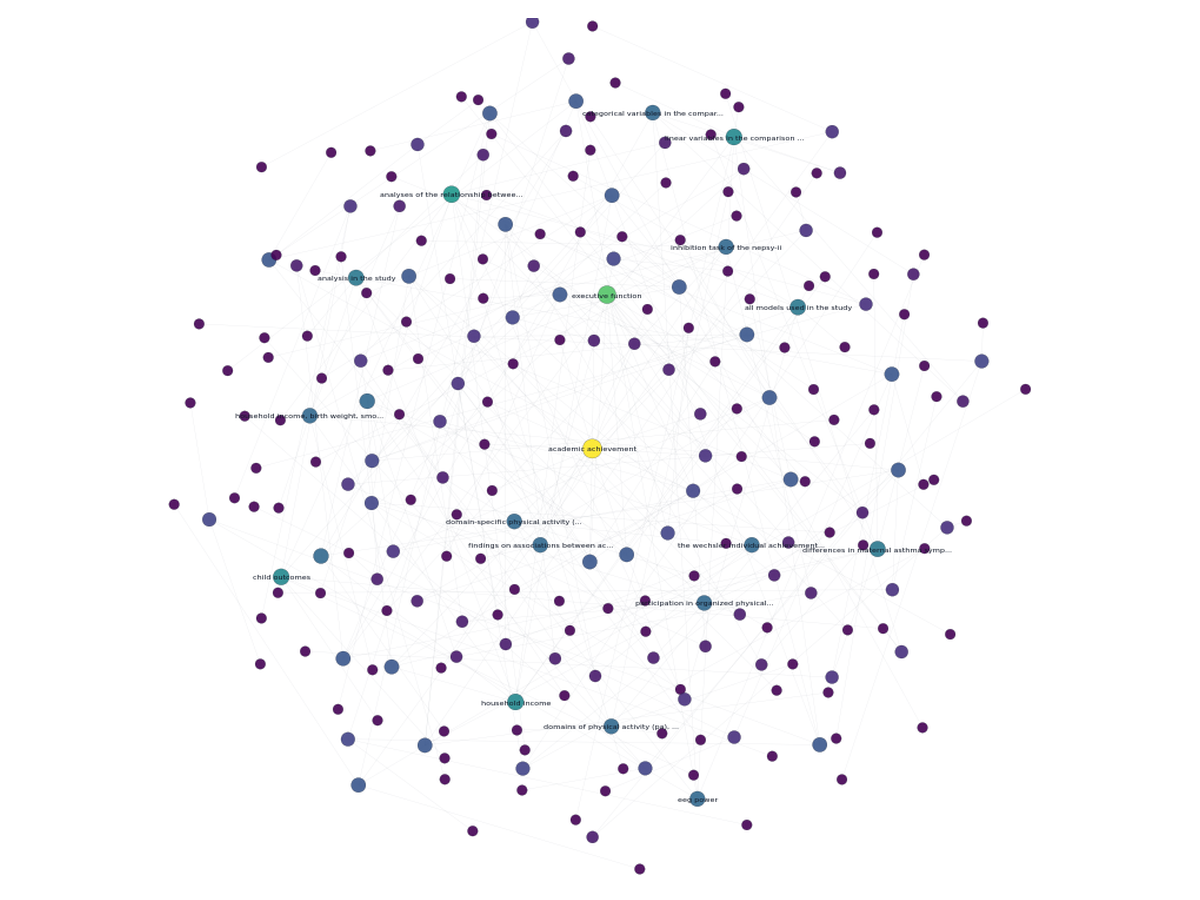}
        \caption{GPT-5.4 full extraction.}
        \label{fig:gpt54-full}
    \end{subfigure}
    \begin{subfigure}[t]{0.24\textwidth}
        \centering
        \includegraphics[width=\textwidth]{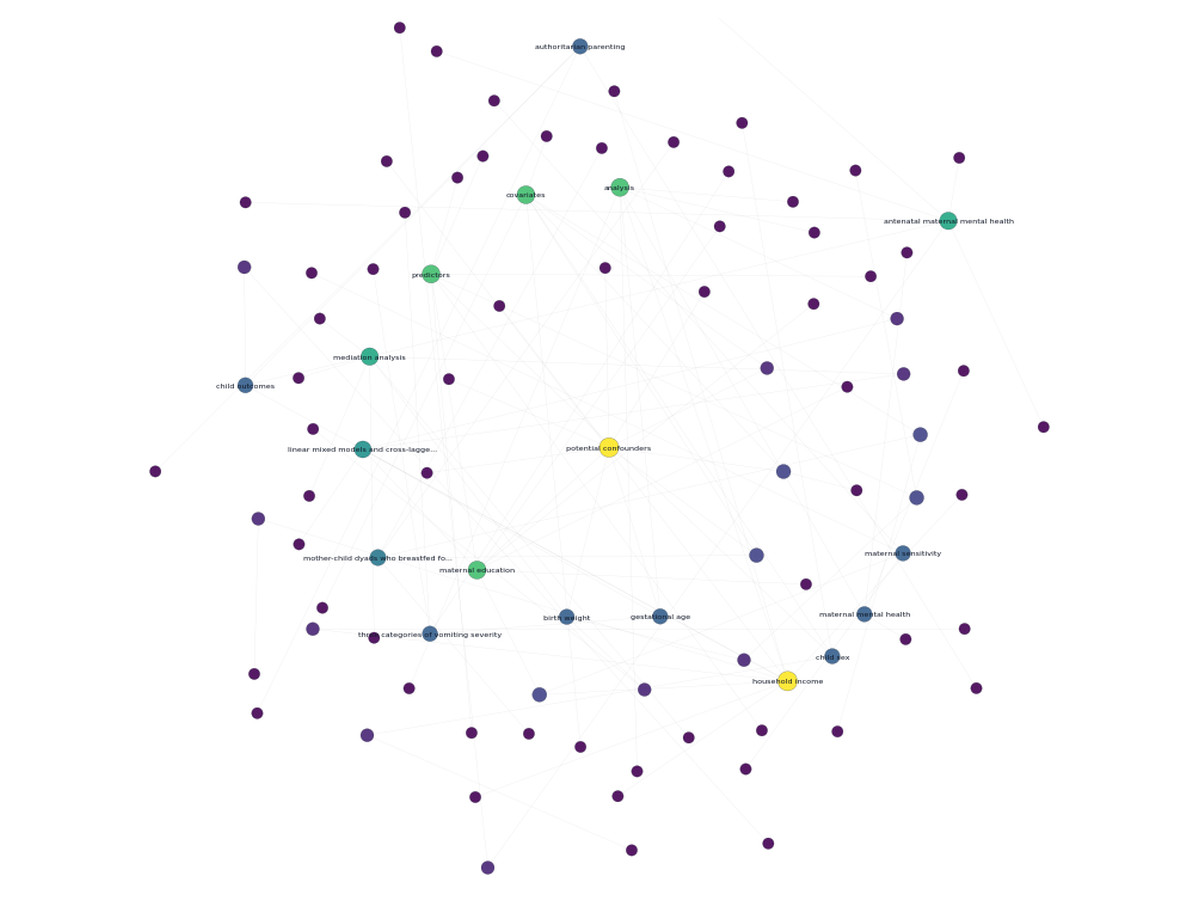}
        \caption{Gemini 3.1 Flash Lite initial extraction.}
        \label{fig:gemini-flash-initial}
    \end{subfigure}
    \hfill
    \begin{subfigure}[t]{0.24\textwidth}
        \centering
        \includegraphics[width=\textwidth]{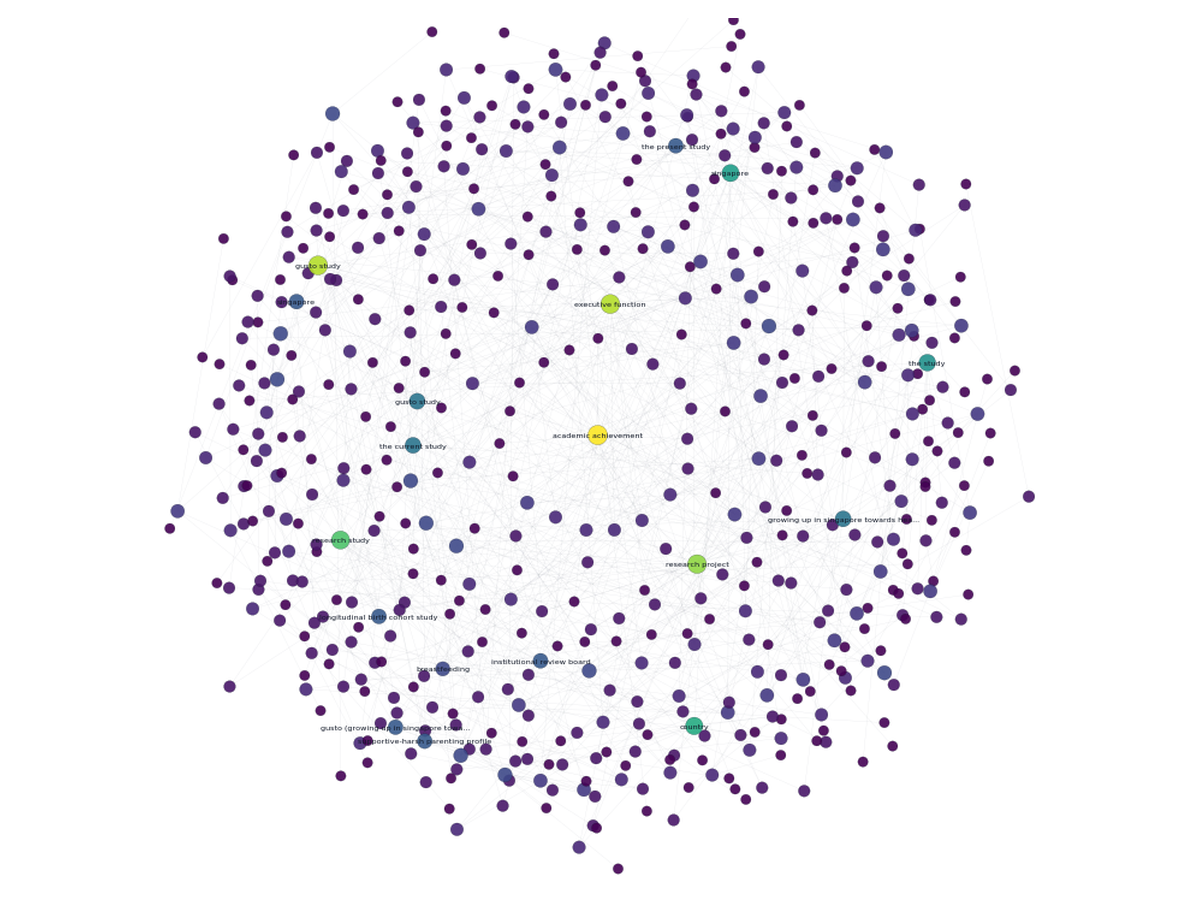}
        \caption{Gemini 3.1 Flash Lite full extraction.}
        \label{fig:gemini-flash-full}
    \end{subfigure}
    \begin{subfigure}[t]{0.24\textwidth}
        \centering
        \includegraphics[width=\textwidth]{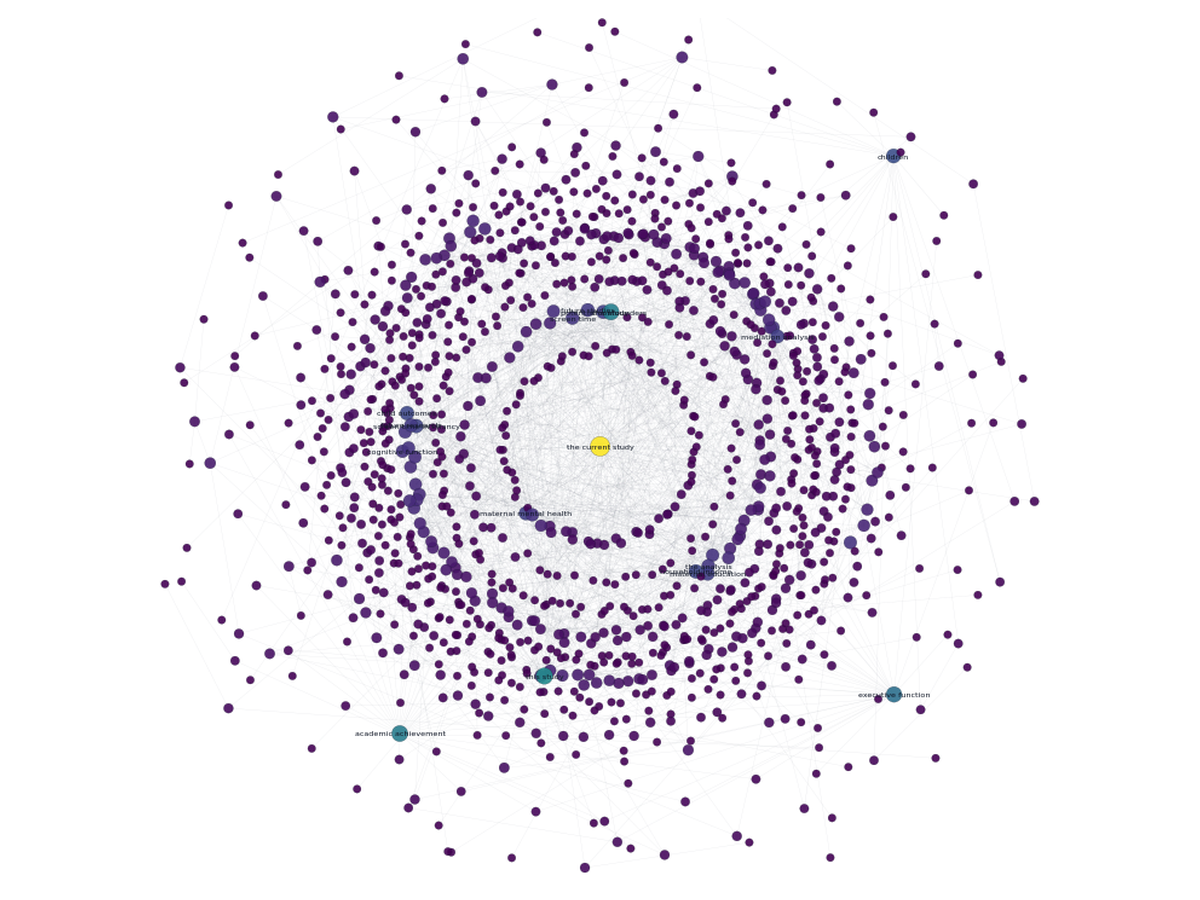}
        \caption{Gemini 3.1 Pro initial extraction.}
        \label{fig:gemini-pro-initial}
    \end{subfigure}
    \hfill
    \begin{subfigure}[t]{0.24\textwidth}
        \centering
        \includegraphics[width=\textwidth]{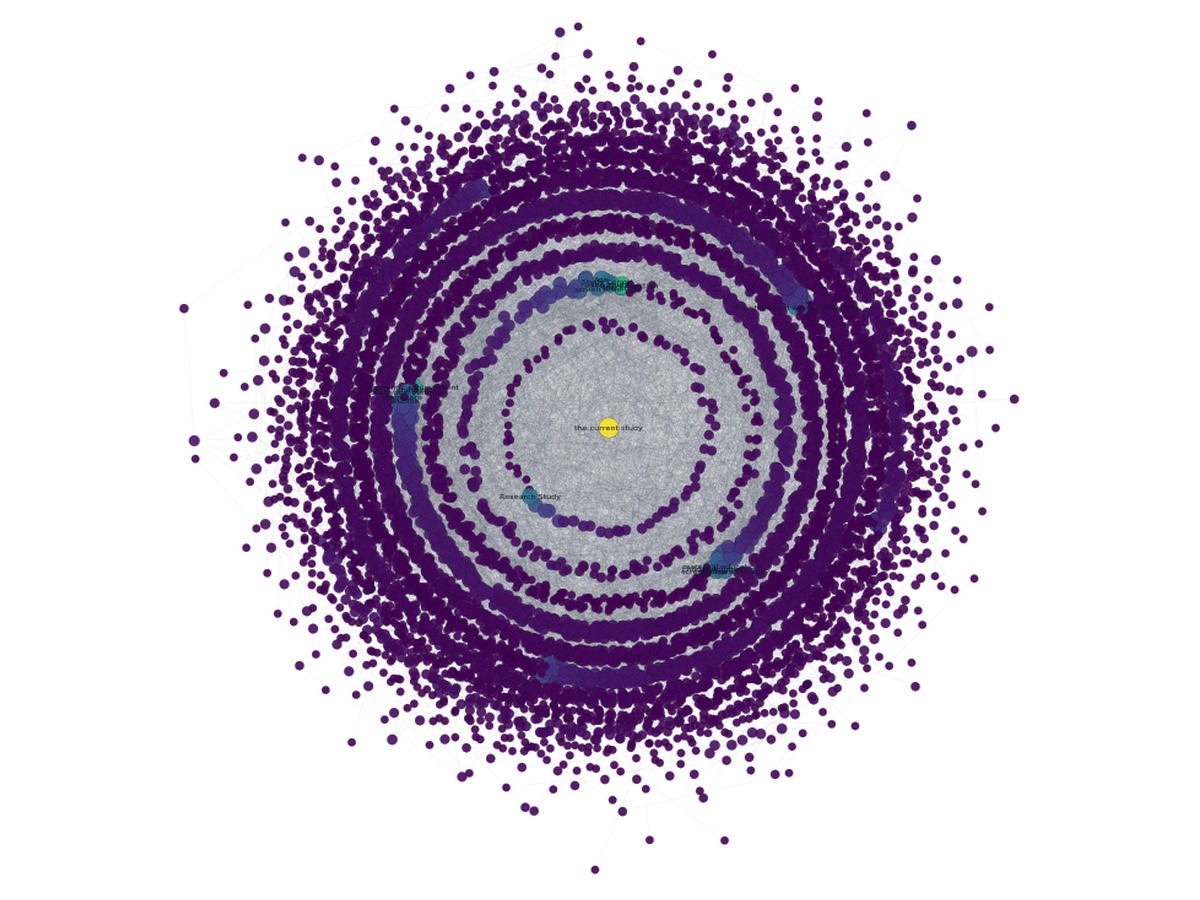}
        \caption{Gemini 3.1 Pro full extraction.}
        \label{fig:gemini-pro-full}
    \end{subfigure}
    \begin{subfigure}[t]{0.24\textwidth}
        \centering
        \includegraphics[width=\textwidth]{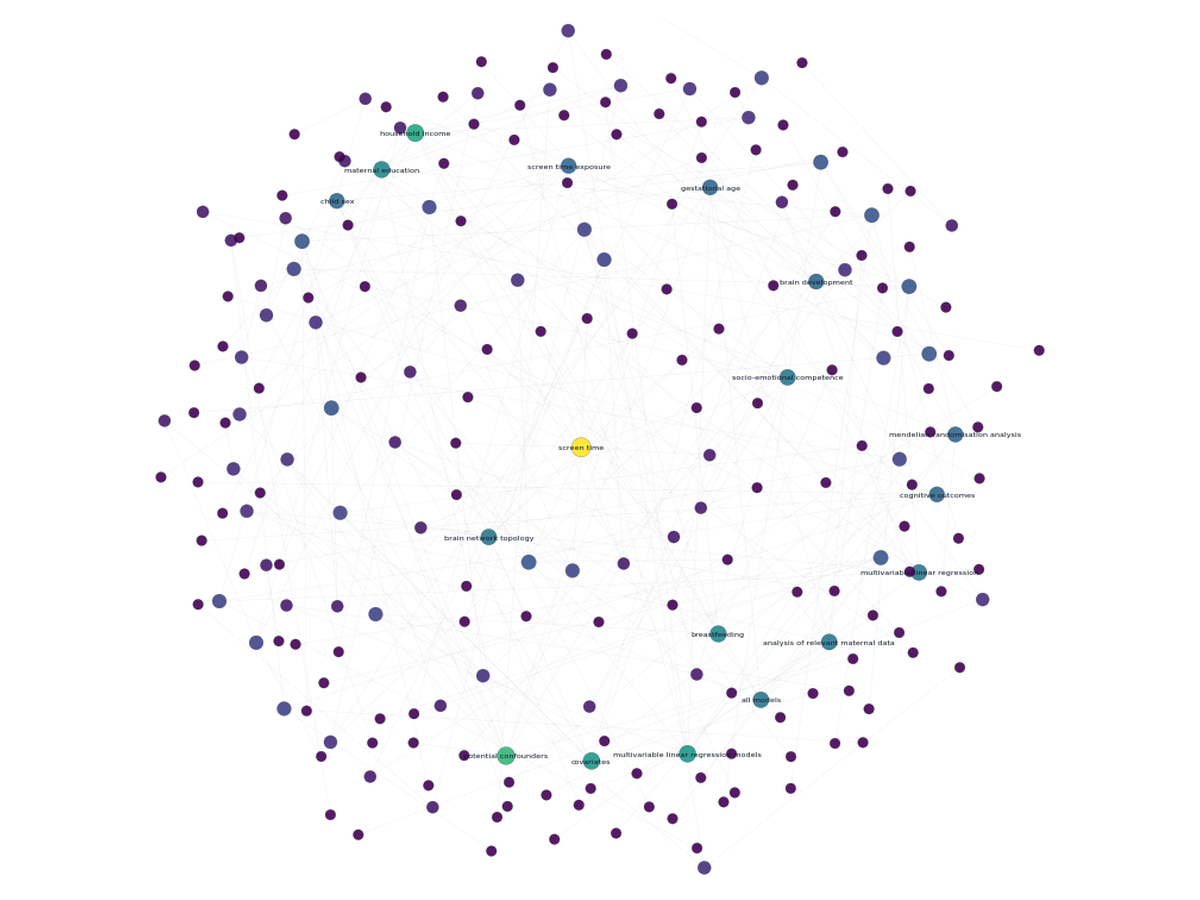}
        \caption{Claude Sonnet 4.6 initial extraction.}
        \label{fig:sonnet-initial}
    \end{subfigure}
    \hfill
    \begin{subfigure}[t]{0.24\textwidth}
        \centering
        \includegraphics[width=\textwidth]{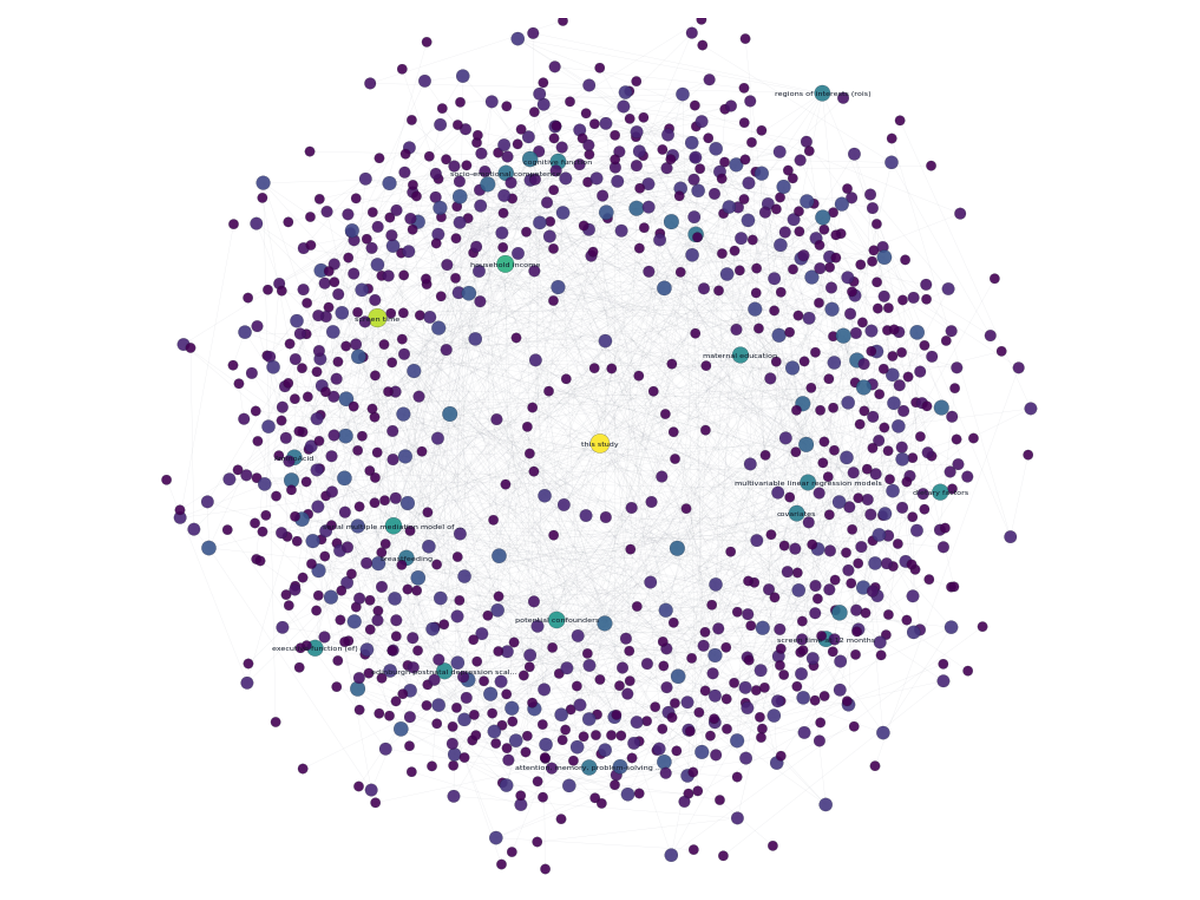}
        \caption{Claude Sonnet 4.6 full extraction.}
        \label{fig:sonnet-full}
    \end{subfigure}
    \begin{subfigure}[t]{0.24\textwidth}
        \centering
        \includegraphics[width=\textwidth]{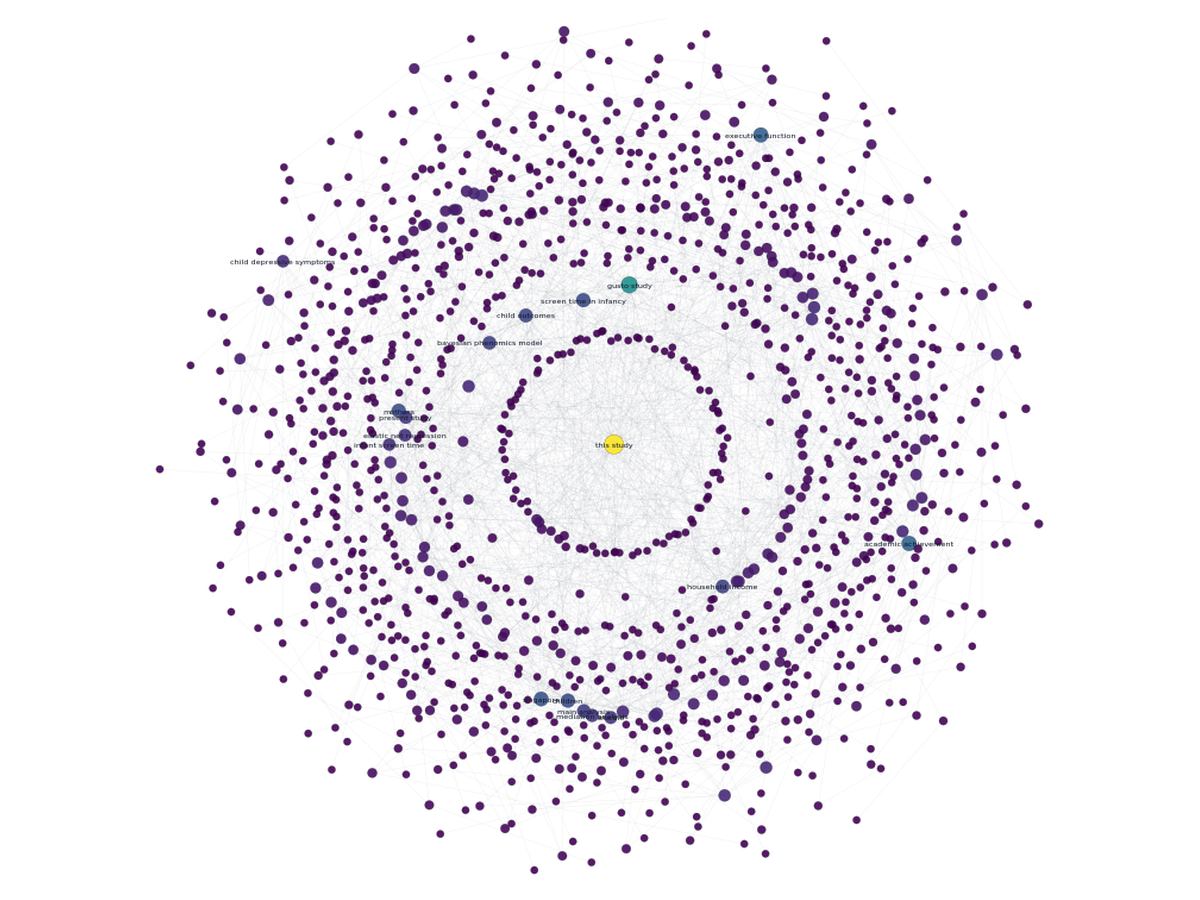}
        \caption{Claude Opus 4.6 initial extraction.}
        \label{fig:opus-initial}
    \end{subfigure}
    \hfill
    \begin{subfigure}[t]{0.24\textwidth}
        \centering
        \includegraphics[width=\textwidth]{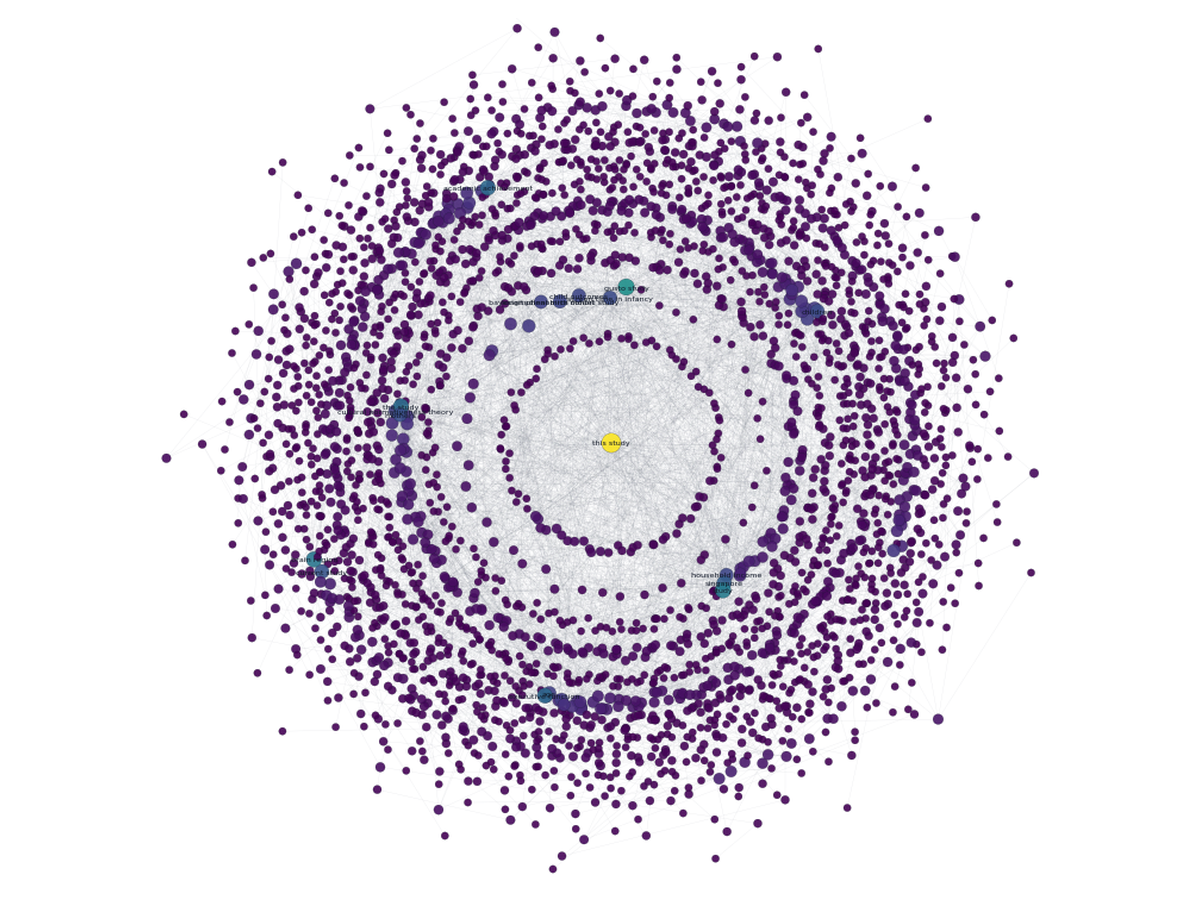}
        \caption{Claude Opus 4.6 full extraction.}
        \label{fig:opus-full}
    \end{subfigure}
    \par\medskip

    \makebox[\textwidth][c]{%
        \begin{subfigure}[t]{0.24\textwidth}
            \centering
            \includegraphics[width=\textwidth]{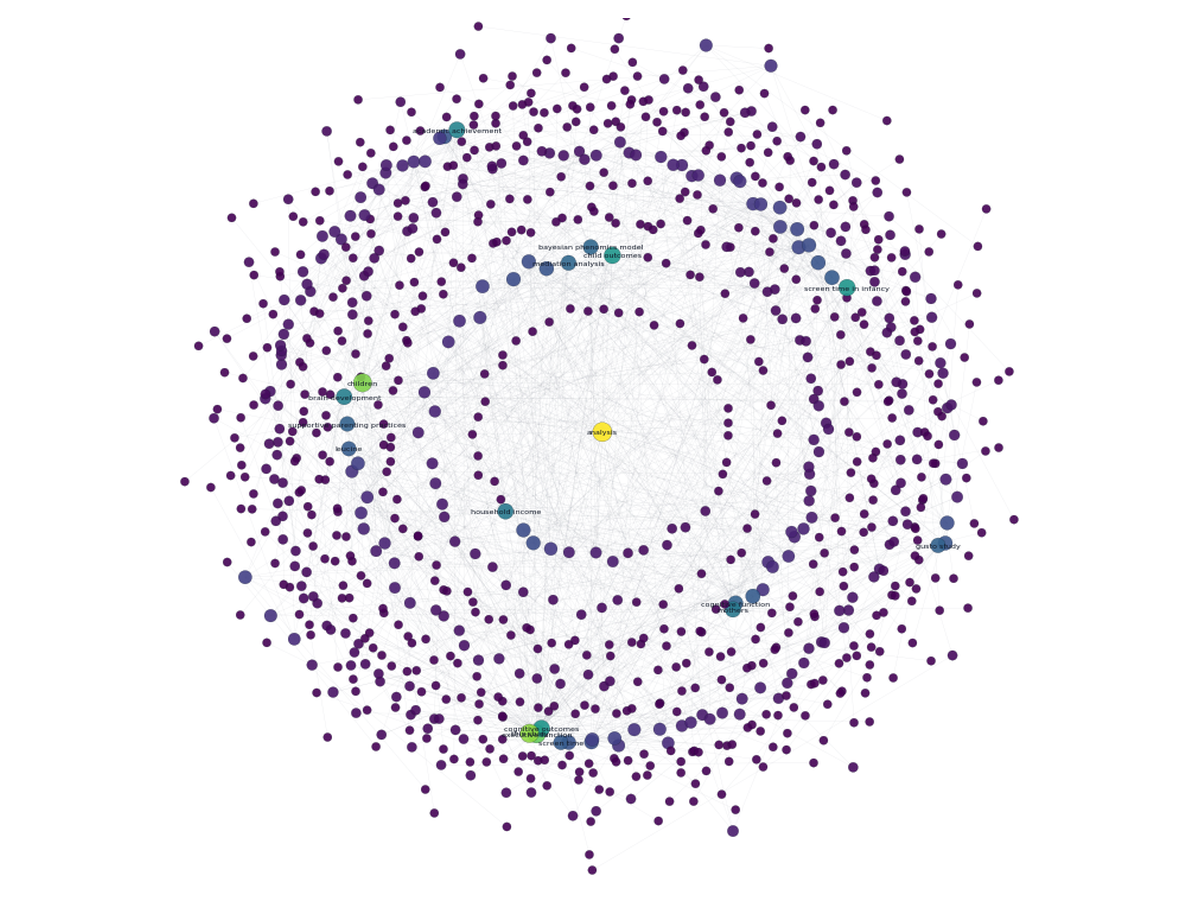}
            \caption{LLaMA 4 Maverick initial extraction.}
            \label{fig:llama4-initial}
        \end{subfigure}
        \begin{subfigure}[t]{0.24\textwidth}
            \centering
            \includegraphics[width=\textwidth]{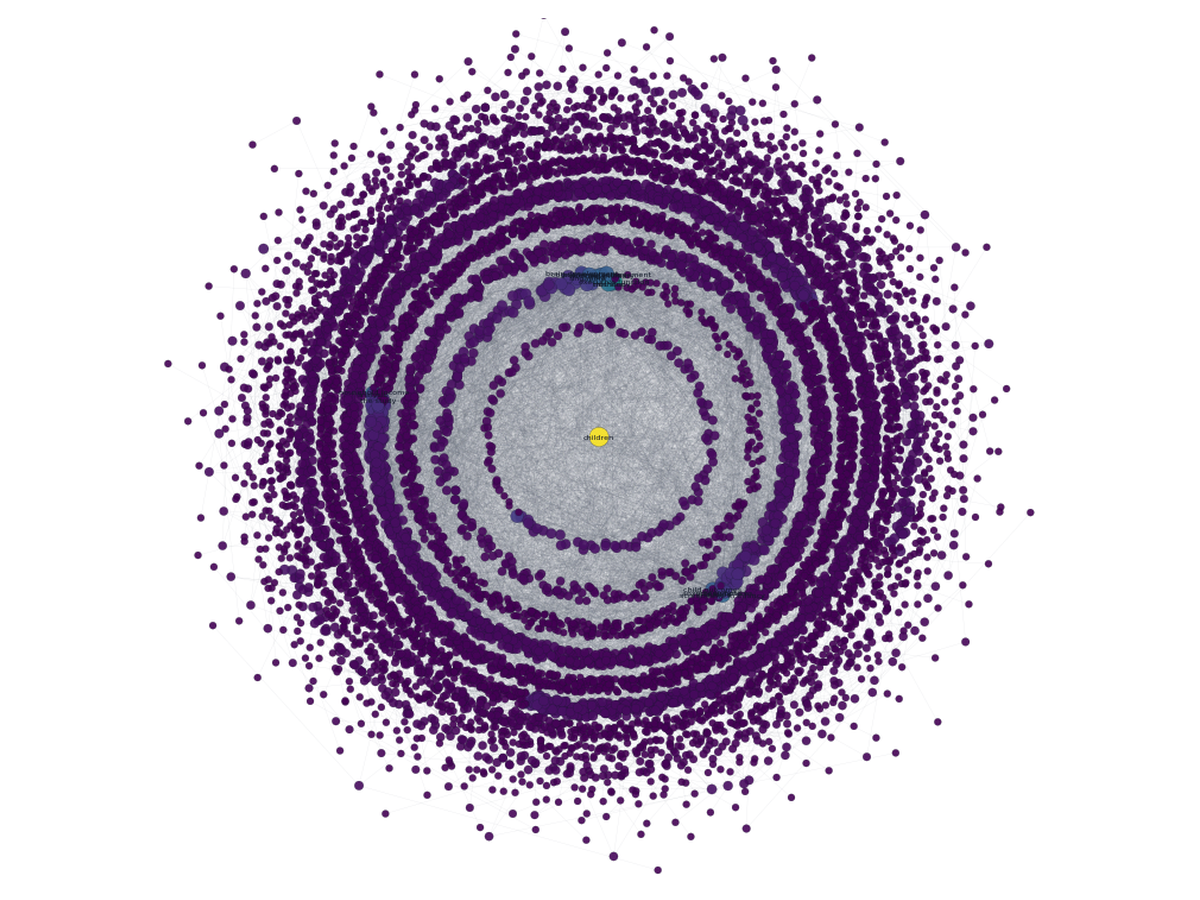}
            \caption{LLaMA 4 Maverick full extraction.}
            \label{fig:llama4-full}
        \end{subfigure}
    }

    \caption{Visualization of knowledge graphs generated by different LLMs after the initial extraction and full construction pipelines.
}
    \label{kg_visualization}
\end{figure*}

Figure~\ref{kg_visualization} provides an overview of the knowledge graphs generated by different LLMs after the initial extraction stage and the full construction pipeline, which includes initial extraction, splitting, and abstraction.

\section{The Necessity of Semantic Evaluation}
We believe that traditional evaluation scripts—such as the WEBNLG evaluation scripts \cite{castro-ferreira-etal-2020-2020}, which compute Precision, Recall, and F1 scores based on token-level matching between predicted and ground truth triples—may not be suitable for accurately assessing the quality of a knowledge graph. Consider the following sentence example from Wiki-NRE dataset \cite{trisedya-etal-2019-neural}. When evaluated using the WEBNLG scripts, the F1 score under the "Partial" criterion is approximately 72.2\%, while the scores under the "Strict" and "Exact" criteria drop to around 61.1\%. However, the output generated by GPT-4 (version GPT-4-1106-preview from Microsoft Azure\footnote{\url{https://learn.microsoft.com/en-us/azure/ai-foundry/openai/concepts/models?tabs=global-standard\%2Cstandard-chat-completions}}) is semantically correct, and the corresponding triples can be directly and accurately derived from the sentence. In fact, the model's output may even surpass the ground truth labels in correctness. For instance, a person holding citizenship in a country is not necessarily born there.
\begin{itemize}
    \item Sentence: Muhammed Ikram is a Pakistani footballer, who is a member of Pakistan national football team.
    \item Ground Truth Labels: [['Muhammed Ikram', 'country of citizenship', 'Pakistan'], ['Muhammed Ikram', 'place of birth', 'Pakistan']]
    \item Structured Triple Generation via GPT-4: [['Muhammed Ikram', 'country of citizenship', 'Pakistan'], ['Muhammed Ikram', 'member of sports team', 'Pakistan national football team']]
\end{itemize}

\section{Downstream Tasks Supported by the Constructed Knowledge Graph}
Currently, our knowledge graph can effectively support domain-specific question answering by providing detailed answers, source citations, and integrated information from multiple sources. Two examples are shown in Figures~\ref{qa_sample_1} and~\ref{qa_sample_2}, we use o4-mini for question answering and Contriever for retrieval, retrieving the 20 and 50 most similar entities, respectively, together with their associated 2-hop triples from the knowledge graph.

As illustrated in Figure~\ref{qa_sample_1}, when combined with information extracted from the tables and figures of the source research papers, our knowledge graph can provide detailed answers together with fine-grained citation information. This makes it easier for domain experts to trace the provenance of the retrieved information and identify potential issues in the original sources. For example, in this case, a typographical inconsistency in the source paper leads to a slight discrepancy in the reported values, i.e., 1.77 versus 1.78.

Figure~\ref{qa_sample_2} further demonstrates that our knowledge graph can answer a question by retrieving and integrating relevant information from multiple sources.

\end{document}